\documentclass[journal]{IEEEtran}
\ifCLASSINFOpdf
  \usepackage[pdftex]{graphicx}
\else
\fi
\usepackage{amsmath}
\newcommand{\dom}{\mathrm{dom}}
\usepackage{amssymb}
\usepackage{multirow}
\usepackage{makecell}
\usepackage{booktabs}
\usepackage{threeparttable}

\usepackage{algorithmic}
\usepackage[caption=false,font=footnotesize]{subfig}
\begin{document}
%
\title{Improving Model Robustness with Latent Distribution Locally and Globally}
%
%
%

\author{Zhuang QIAN,
        Shufei ZHANG,
        Kaizhu HUANG,
        Qiufeng WANG,
        Rui ZHANG,
        Xinping YI
\IEEEcompsocitemizethanks{

\IEEEcompsocthanksitem Zhuang QIAN and Shufei Zhang contributed equally to this work.
\IEEEcompsocthanksitem Kaizhu HUANG (Corresponding Author) is with Data Science Research Center, Duke Kunshan University. E-mail:\{Kaizhu.Huang\}@dukekunshan.edu.cn
\IEEEcompsocthanksitem Zhuang QIAN and Xinping YI are with Department of Electrical Engineering and Electronics, University of Liverpool. E-mail:\{Zhuang.Qian, Xinping.Yi\}@liverpool.ac.uk
\IEEEcompsocthanksitem Shufei ZHANG is with Shanghai Artificial Intelligence Laboratory. E-mail:\{zhangshufei\}@@pjlab.org.cn
\IEEEcompsocthanksitem Qiufeng WANG is with School of Advanced Technology, Xi'an Jiaotong-Liverpool University.  E-mail:{qiufeng.wang}@xjtlu.edu.cn
\IEEEcompsocthanksitem Rui ZHANG is with Department of Foundational Mathematics, Xi'an Jiaotong-Liverpool University. E-mail:{rui.zhang02}@xjtlu.edu.cn}

}

%
%

\markboth{Journal of \LaTeX\ Class Files,~Vol.~14, No.~8, August~2015}%
{Shell \MakeLowercase{\textit{et al.}}: Bare Demo of IEEEtran.cls for IEEE Journals}
%



\maketitle

\begin{abstract}
In this work, we consider model robustness of deep neural networks against adversarial attacks from a global manifold perspective. 
Leveraging both the local and global latent information, we propose a novel adversarial training method through robust optimization, and a tractable way to generate Latent Manifold Adversarial Examples (LMAEs) via an adversarial game between a discriminator and a classifier.
The proposed adversarial training with latent distribution (ATLD) method defends against adversarial attacks by crafting LMAEs with the latent manifold  in an unsupervised manner.
ATLD preserves the local and global information of the latent manifold and promises improved robustness against adversarial attacks.
To verify the effectiveness of our proposed method, we conduct extensive experiments over different datasets (e.g., CIFAR-10, CIFAR-100, SVHN) with different adversarial attacks (e.g., PGD, CW), and show that our method substantially outperforms the state-of-the-art (e.g., Feature Scattering) in adversarial robustness by a large accuracy margin. The source codes are available at https://github.com/LitterQ/ATLD-pytorch.
\end{abstract}


%
\IEEEpeerreviewmaketitle

\section{Introduction}
Whilst Deep Neural Networks (DNNs) have achieved tremendous success in the past decade, the model robustness emerges as one of the greatest challenges in safety-critical applications (e.g., self-driving, healthcare), as it appears DNNs
can be easily fooled by adversarial examples or perturbations of the input data~\cite{Deeplearning,resnet,lstm}. 
Adversarial examples are
ubiquitous in a variety of tasks such as image classification~\cite{FSGM}, segmentation~\cite{ad_seg}, and speech recognition~\cite{ad_speech}, text classification~\cite{yang2020greedy}
~and therefore raise great concerns with the robustness of learning models and have drawn enormous attention. 

To defend against adversarial examples, great efforts have been made to improve the model robustness~\cite{Adversarial_logit_pairing,Adversarial_noise_layer,Bilateral,feature_scattering}. 
While adversarial attacks and defences turn out to be an arms race, adversarial training has been shown as one of the most promising techniques, for which the models are trained 
with adversarially-perturbed samples rather than clean data~\cite{FSGM,PGD,lyu}. 
The adversarial training 
can be formulated as a minimax robust optimization problem, which can be seen as
a min-max game between the adversarial perturbations and classifier training. Namely, the indistinguishable adversarial perturbations are crafted to maximize the chance of misclassification, while the classifier is trained to minimize the loss due to such perturbed data.
In the literature, the majority of adversarial perturbations are crafted with data labels 
in a supervised manner (e.g.,~\cite{PGD}).  

\begin{figure*}[htbp]
	\centering
    \subfloat[Original Data]{
		\includegraphics[width=0.17\textwidth]{{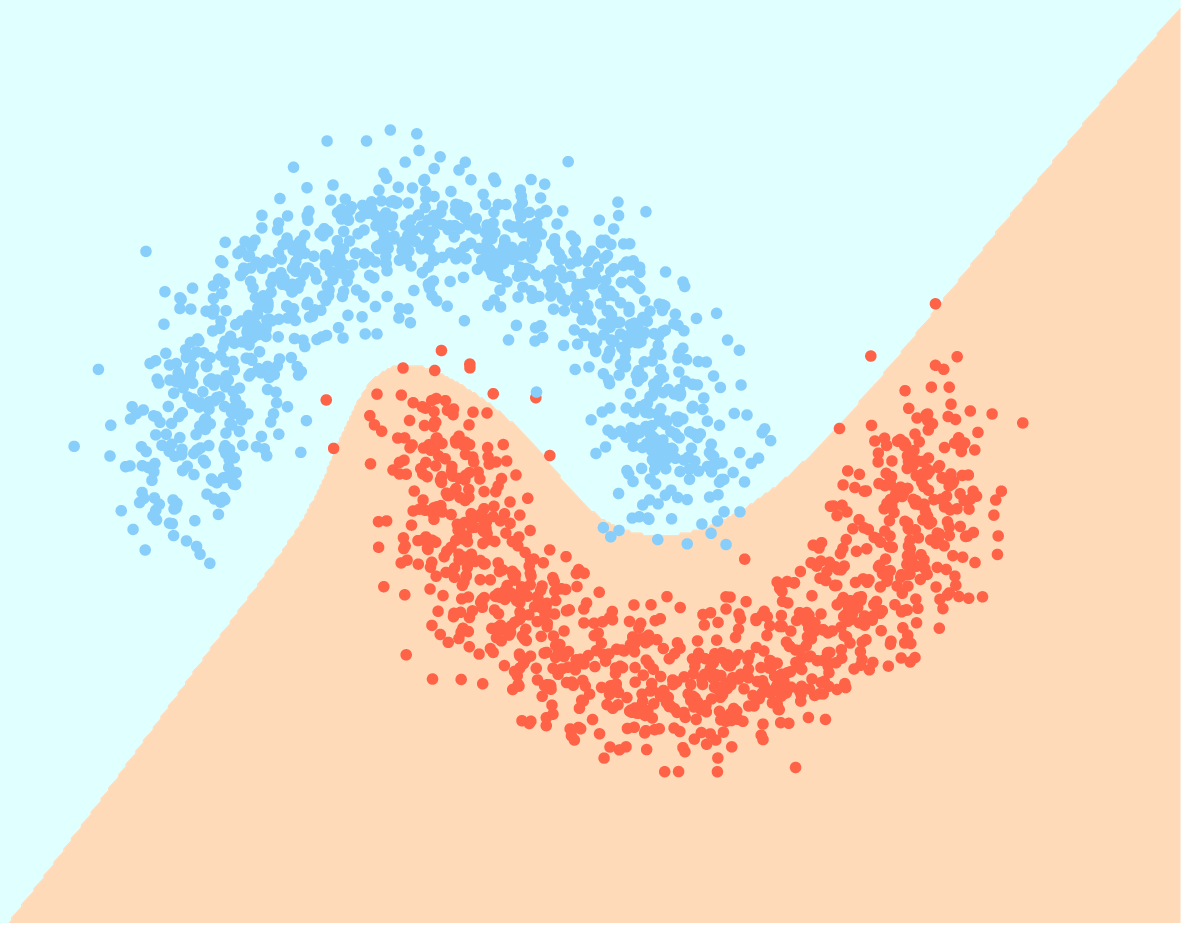}}
		\label{subfig:toy1_org}
		}
	\subfloat[PGD]{
		\includegraphics[width=0.17\textwidth]{{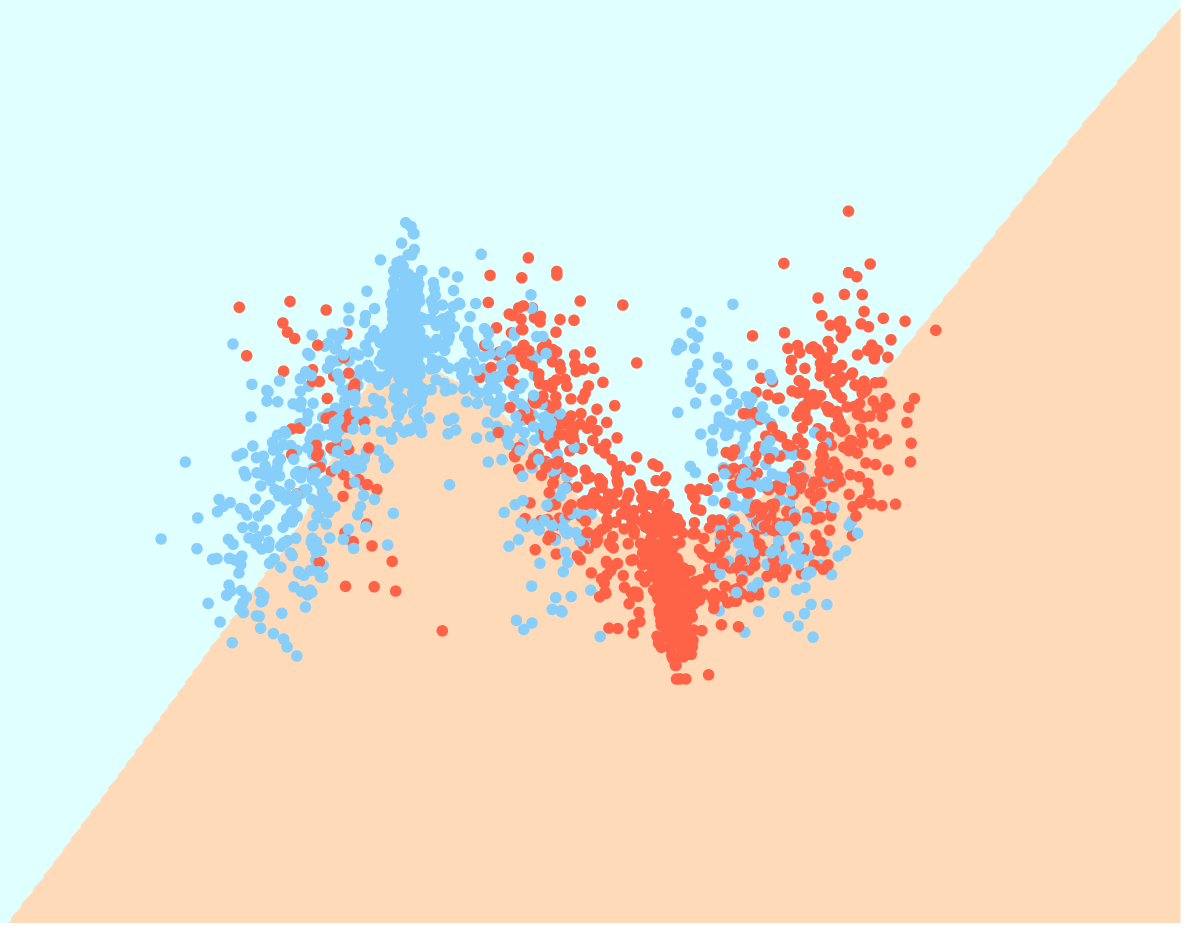}}
		\label{subfig:toy1_pgd}
	}
 	\subfloat[TRADES]{
		\includegraphics[width=0.17\textwidth]{{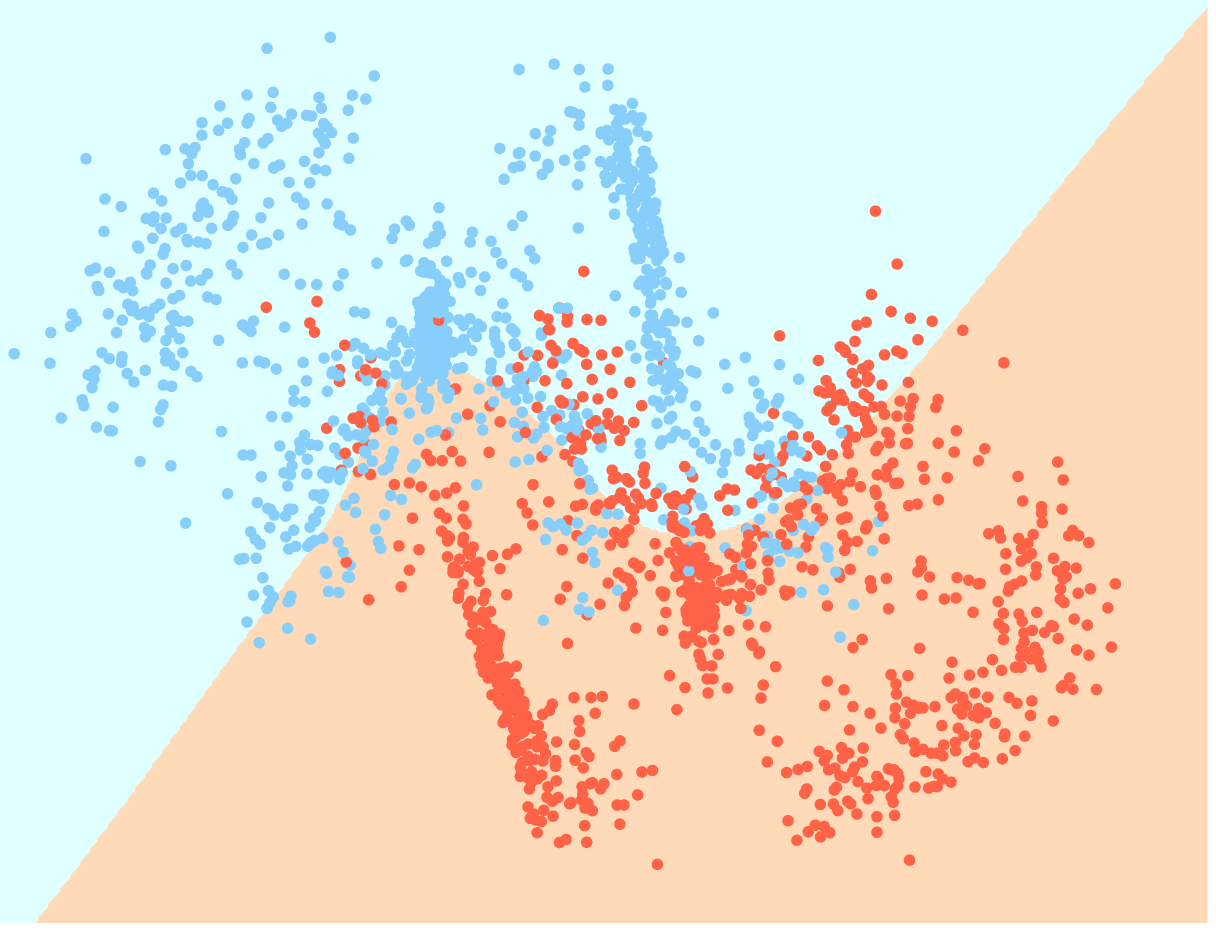}}
		\label{subfig:toy1_trades}
	}
	\subfloat[Feature Scattering]{
		\includegraphics[width=0.17\textwidth]{{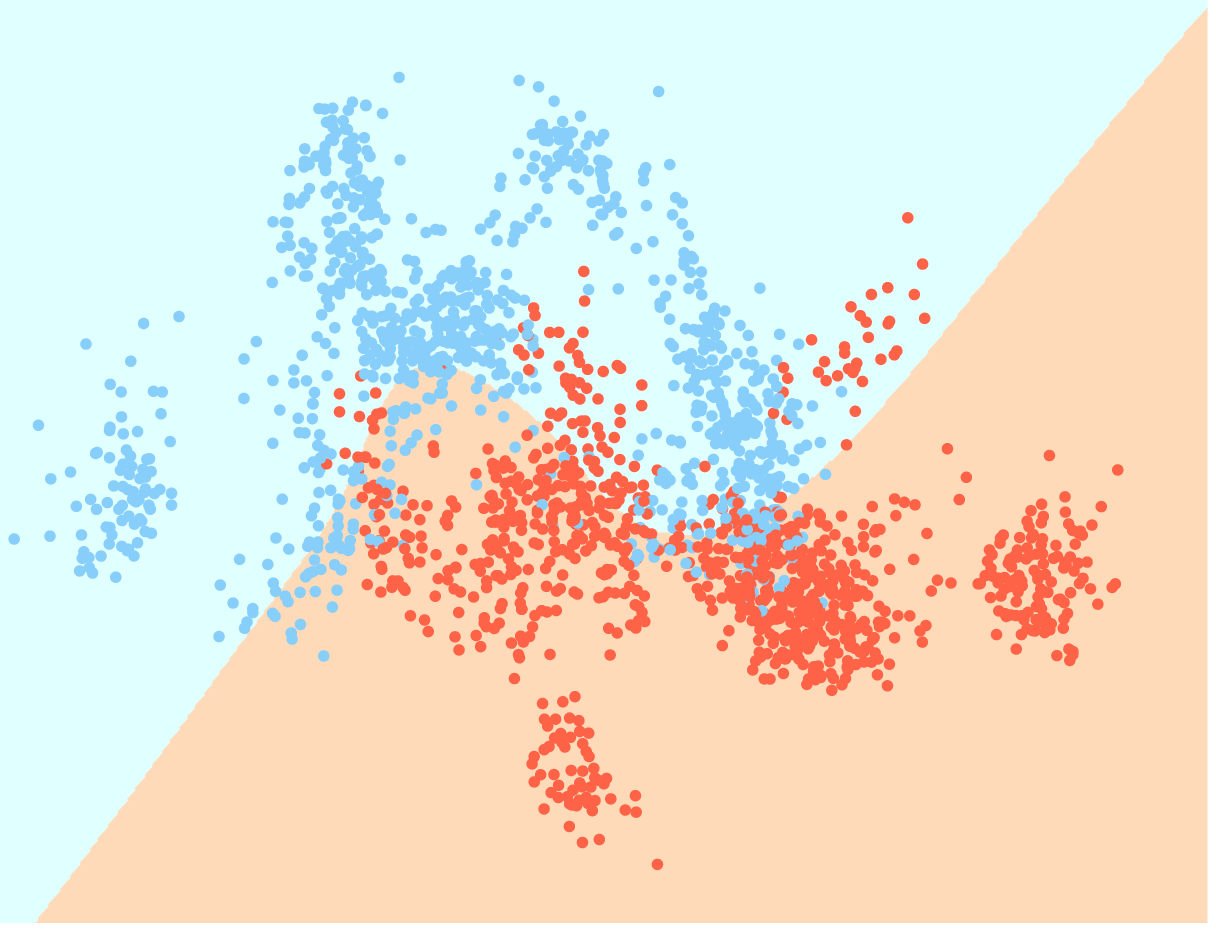}}
		\label{subfig:toy1_fs}
	}
	\subfloat[Ours]{
		\includegraphics[width=0.17\textwidth]{{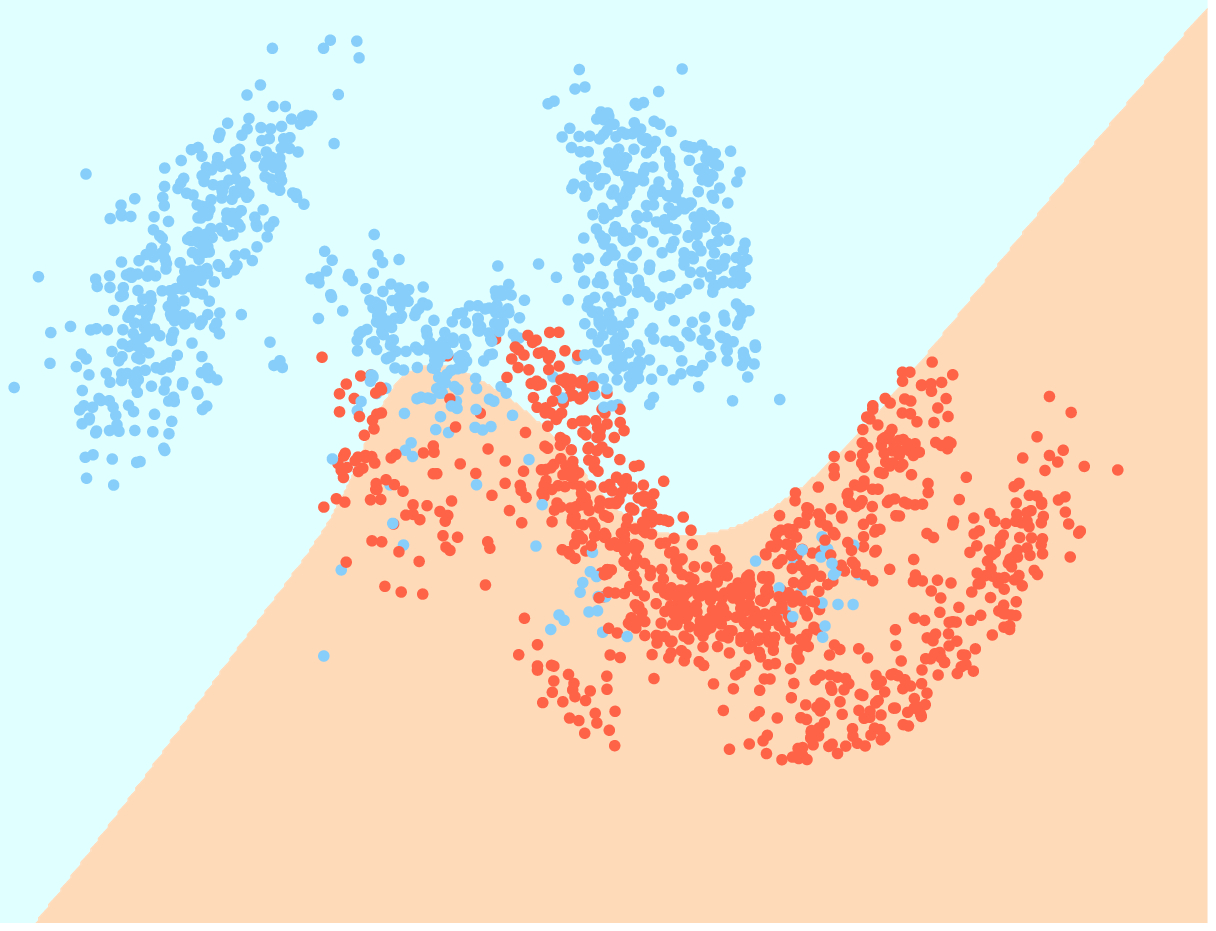}}
		\label{subfig:toy1_ours}
	}\\ \vspace{-5pt}
			\subfloat[Original]{
		\includegraphics[width=0.17\textwidth]{{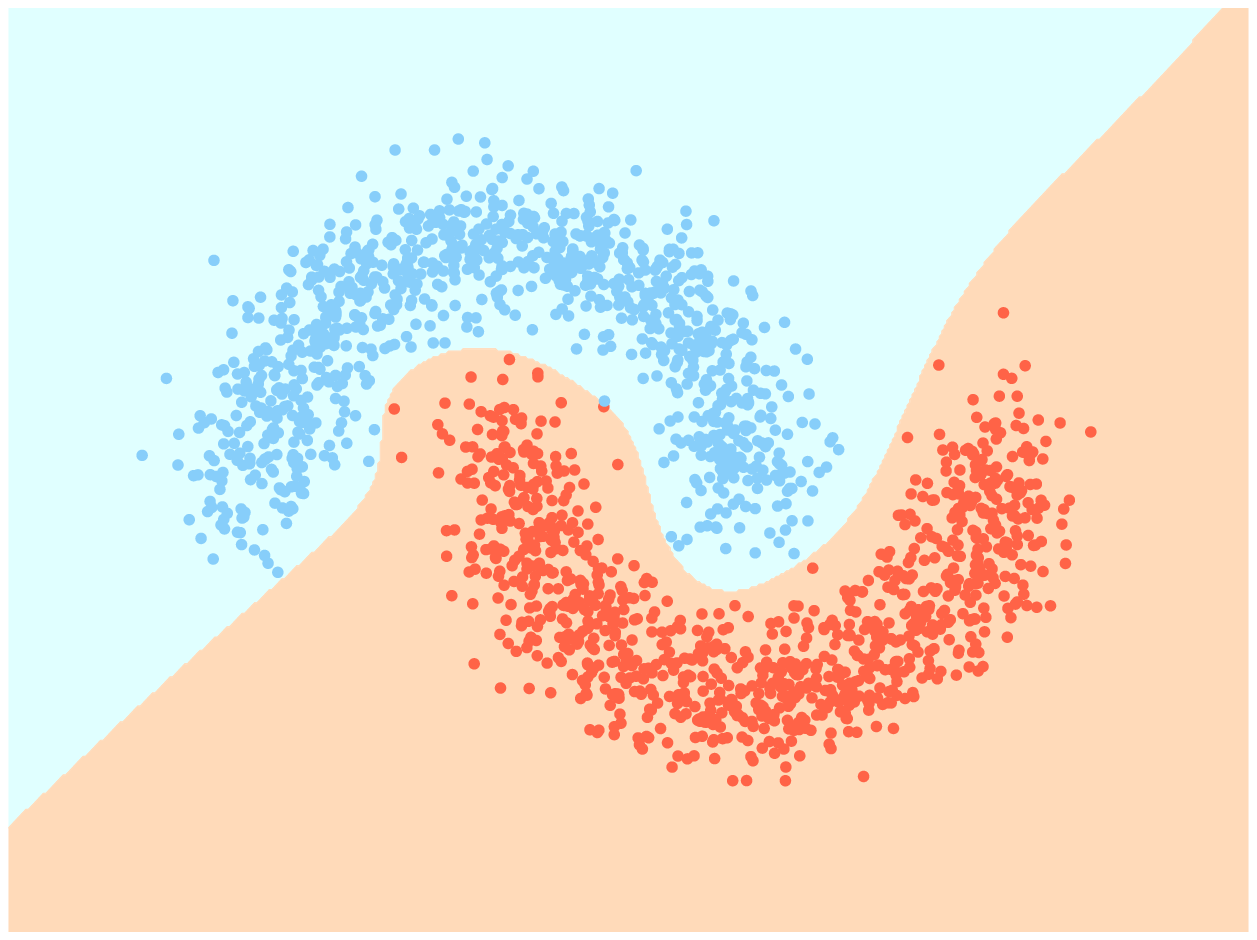}}
		\label{subfig:toy3_org}
	}
	\subfloat[PGD-AT]{
		\includegraphics[width=0.17\textwidth]{{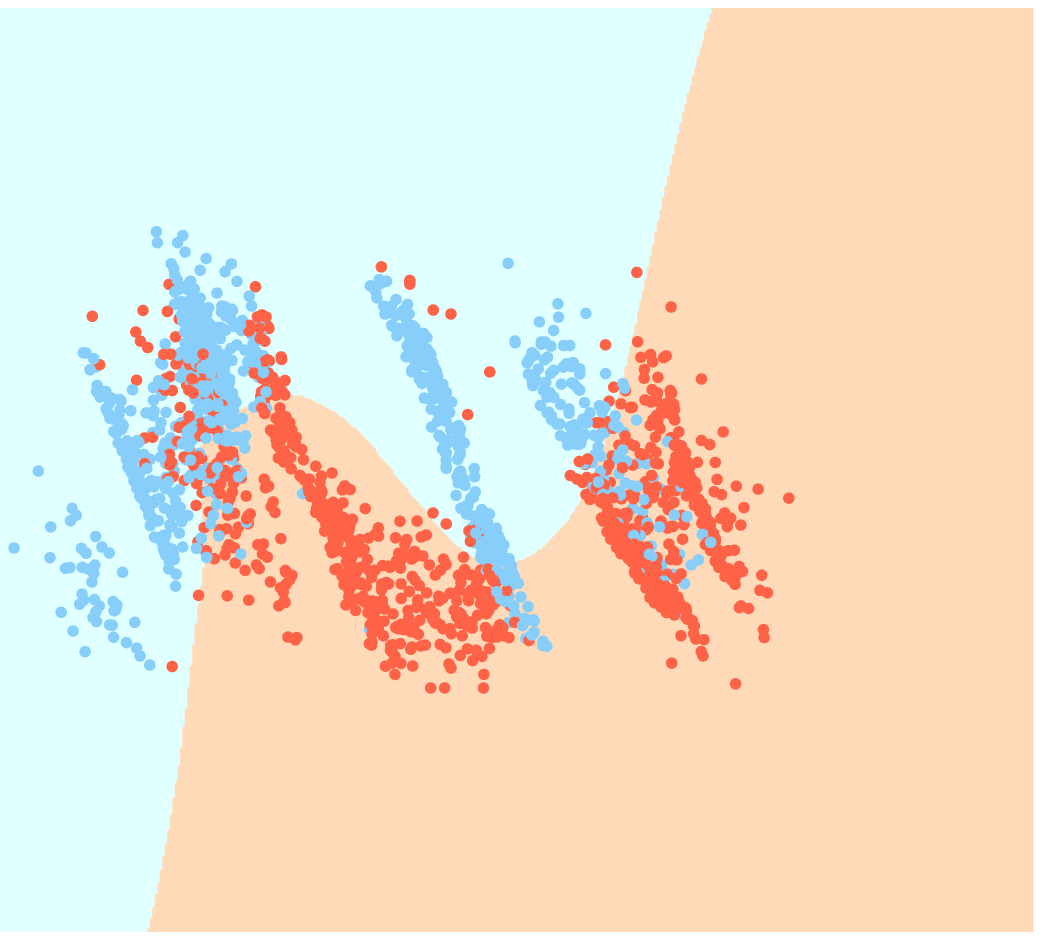}}
		\label{subfig:toy3_pgd}
	}
 	\subfloat[TRADES]{
		\includegraphics[width=0.17\textwidth]{{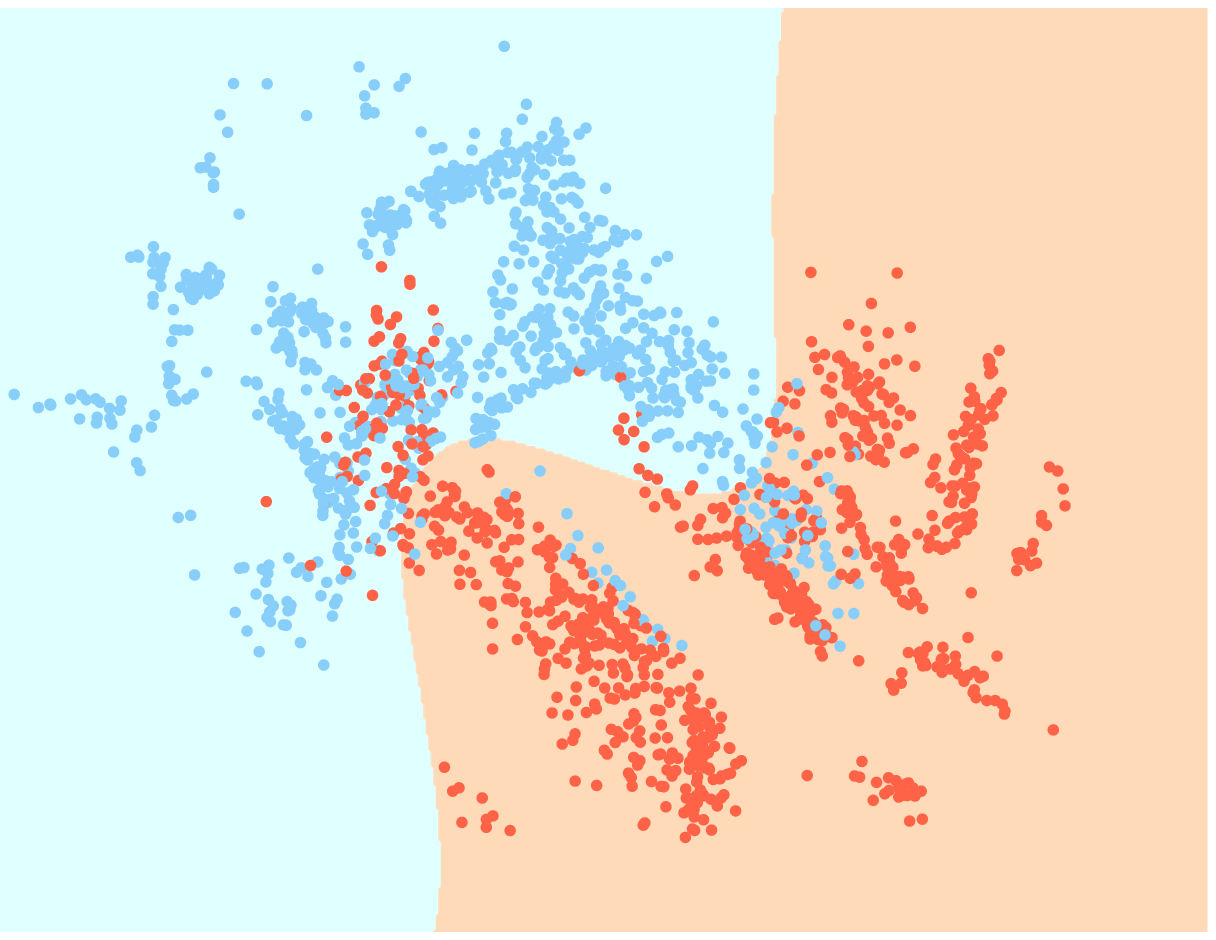}}
		\label{subfig:toy3_trades}
	}
	\subfloat[Feature Scattering]{
		\includegraphics[width=0.17\textwidth]{{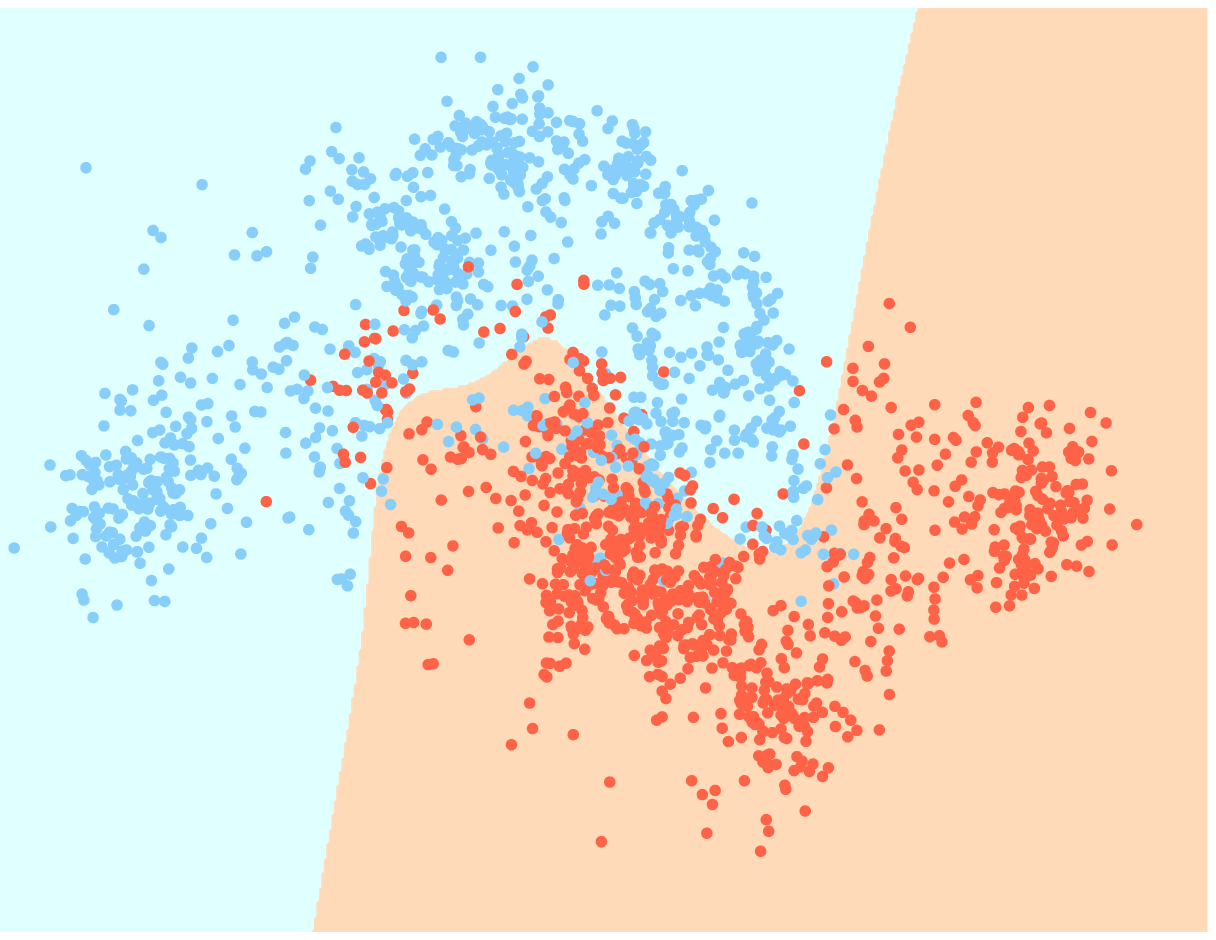}}
		\label{subfig:toy3_fs}
	}
	\subfloat[Ours]{
		\includegraphics[width=0.17\textwidth]{{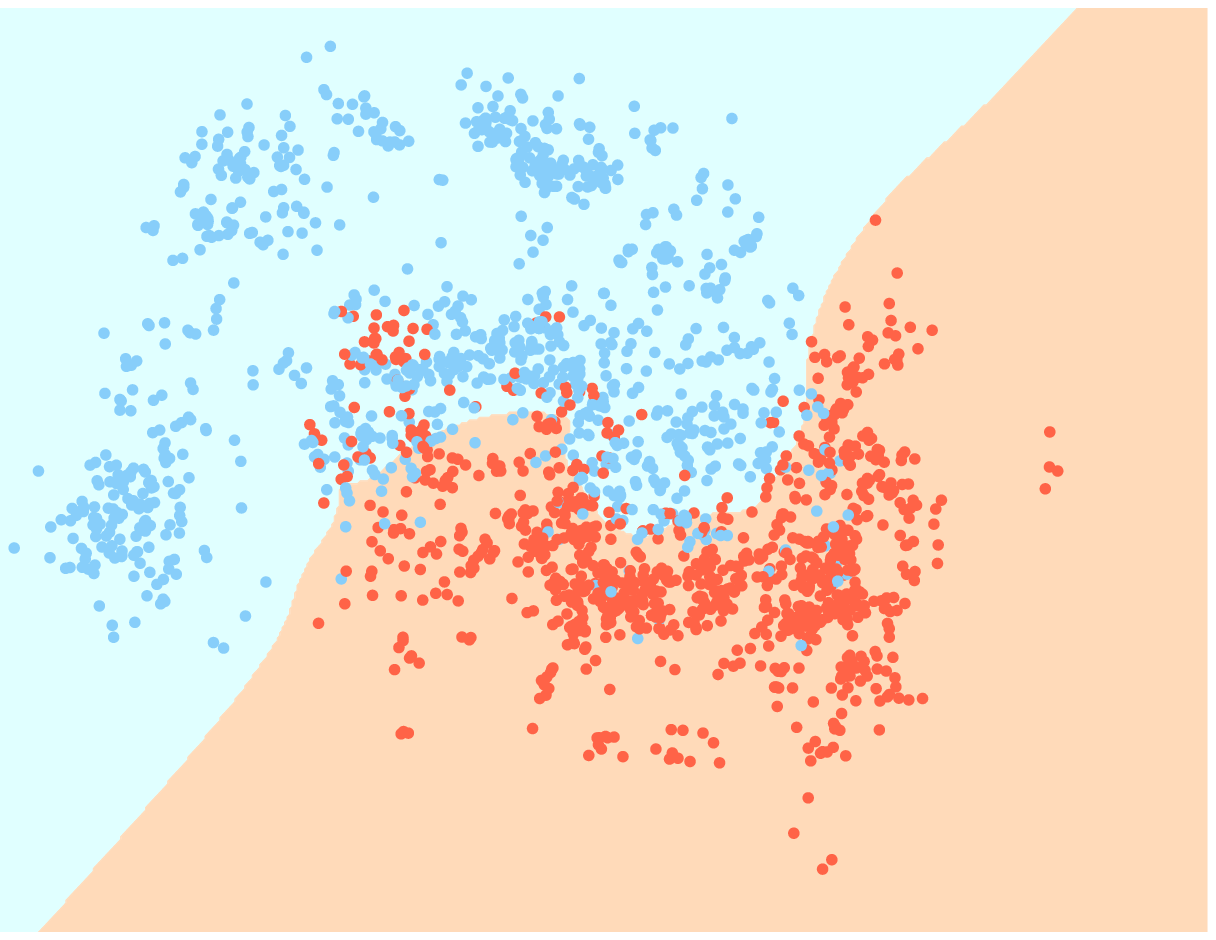}}
		\label{subfig:toy3_ours}
	}
	\label{fig:toy1}
	\caption{
	An illustrative example of data manifold and decision boundary after different adversarial perturbation schemes (Top) and adversarial training (Bottom). Top: (a) Original data without perturbation; perturbed data using (b) PGD, a supervised adversarial attack method; (c) TRADES attack method; (d) Feature Scattering, and (e) the proposed method with LMAEs. The overlaid boundary is from the model trained on clean data. Bottom: (f) Standard training with original data; (g) adversarial training (AT) with PGD perturbed data; (h) TRADES training; (i) Feature Scattering, and (j) the proposed ATLD method. The overlaid decision boundary after the various adversarial training is applied.
	}
\end{figure*}

As a matter of fact,
these adversarial training approaches are 
restrictive, as the perturbations are produced individually within the local region and in a supervised way -- the global data manifold information is unexplored --
such global information however proves crucial for attaining better robustness and generalization performance. 
It is worth noting that 
the locally and individually generated adversarial examples may corrupt the underlying data structure and would be typically biased toward the decision boundary. Therefore, the features inherent to the data distribution might be lost, which  limits the robustness 
even if adversarial training is applied~\cite{ad_bug,ad_generalization}. To illustrate this point, we show a toy example in Figure~\ref{fig:toy1}. As clearly observed, adversarially-perturbed examples generated by PGD, one of the most successful adversarial attacks, corrupt the data manifold, 
inevitably leading to worse robustness performance when adversarial training aims to defend against such perturbed examples. 
Further, although the recent proposed Feature Scattering in~\cite{feature_scattering} can partially alleviate this issue, the corruptions on the data manifold do still exist.

To address the above limitation, we first propose a novel method to generate adversarial examples, namely \textit{Latent Manifold Adversarial Examples (LMAEs)}, for training. The LMAEs are produced from the input space to `deceive' the global latent manifold rather than to fool the loss function as in traditional adversarial training. Instead of maximizing the loss function of the classifier, LMAEs aim to maximize the divergence between the latent distributions of clean data and their adversarial counterparts. Moreover, since the label information is not required when crafting the LMAEs, the resulting LMAEs would not be biased toward the decision boundary, which can be clearly observed in Figure~\ref{fig:toy1} (e) and (j). Further, we propose a novel adversarial training method -- Adversarial Training with Latent Distribution (ATLD) -- which considers the data distribution globally by training the classifier against LMAEs. With ATLD, the global information of the latent data manifold could be well preserved, so that the classifier could generalize well on the unseen adversarial examples, thus leading to better model robustness. 

Nevertheless, evaluating the divergence between the latent distributions of clean data and adversarial examples is still challenging. We leverage a discriminator to estimate the divergence between two distributions to make it tractable. We reformulate the proposed ATLD as \textit{a minimax game between a discriminator and a classifier for the LMAEs}. The LMAEs are crafted according to the loss of the discriminator to make different implicitly 
the latent distributions of the clean and perturbed data, while the classifier is trained to decrease the discrepancy between these two latent distributions and promote accurate classification on the LMAEs as Figure~\ref{fig:framework} shows. 

On the empirical front, first, we illustrate in Figure~\ref{fig:toy1} that our proposed method can preserve more information about the original distribution and learn a better decision boundary than the existing adversarial training methods. The figure shows that both the AT, TRADES and the FS alter the original decision boundary significantly. Moreover, it can be observed that the adversarial training with PGD and TRADES corrupts the data manifold completely. On the other hand, FS appears able to partially retain the data manifold information since it  considers 
the inter-sample relationship locally. Nonetheless, its decision boundary appears non-smooth, which  may degrade the performance. In contrast, our proposed method considers retaining the global information of the data manifold and thus obtaining a nearly optimal decision boundary. This may explain why our proposed ATLD method could outperform the other approaches.
Furthermore, in order to enhance ATLD during the inference procedure, we propose a novel inference method named Inference with Manifold Transformation (IMT) to generate the specific perturbations through the discriminator network to diminish the impact of the adversarial attack as detailed discussion in Section~\ref{sec:IMT}.

We test our method on three different datasets: CIFAR-10, CIFAR-100 and SVHN with the most commonly used PGD, CW and FGSM attacks. Our ATLD method outperforms the state-of-the-art methods by a large margin. e.g. ATLD improves over Feature Scattering~\cite{feature_scattering} by $17.0\%$ and $18.1\%$ on SVHN for PGD20 and CW20 attacks. Our method is also superior to the  conventional adversarial training method~\cite{PGD}, boosting the performance by $32.0\%$ and $30.7\%$ on SVHN for PGD20 and CW20 attacks. When evaluating ATLD against more recent and stronger attacks such as Autoattack~\cite{autoattack} and Rays~\cite{rays}, ATLD also outperforms other state-of-the-art competitors.

\section{Related Work}
\textbf{Adversarial Training.} 
    Adversarial training is a family of techniques to improve the model robustness~\cite{PGD,lyu,zhai2022adversarial,yang2022toward}. It trains the DNNs with adversarially-perturbed samples instead of clean data. Some approaches extend the conventional adversarial training by injecting the adversarial noise into hidden layers to boost the robustness of latent space~\cite{ad_feature,Adversarial_noise_layer,apply_latent,Noise_Propagation}. All of these approaches \textit{generate the adversarial examples by maximizing the loss function with the label information}. However, the structure of the data distribution is destroyed since the perturbed samples could be highly biased toward the non-optimal decision boundary~\cite{feature_scattering}. The poor robust generalization of adversarial training is attributed to the significantly dispersed latent representations generated by training and testing adversarial data~\cite{scr}. Our proposed method has a similar training scheme to adversarial training by replacing clean data with perturbed ones. Nevertheless, our method generates the adversarial perturbations without the label information which weakens the impact of non-optimal decision boundary and can retain more information on the underlying data distribution. 
    
 \textbf{Manifold-based Adversarial Training.}
\cite{pixel_defend} proposes to generate the adversarial examples by projecting on a proper manifold.~\cite{feature_scattering} leverages the manifold information in the form of an inter-sample relationship within the batch to generate adversarial perturbations. Virtual Adversarial Training and Manifold Adversarial Training are proposed to improve model generalization and robustness against adversarial examples by ensuring the local smoothness of the data distribution~\cite{MAT2018,miyato2017virtual}. Some methods are designed to enforce the local smoothness around the natural examples by penalizing the difference between the outputs of adversarial examples and clean counterparts~\cite{Adversarial_logit_pairing,Jacobian_Adv,Jacobia_ECCV}. All of these methods just \textit{leverage the local information} of the distribution or manifold. Differently, our method generates the perturbations additionally considering the structure of distribution globally.

 \textbf{Unsupervised Domain Adversarial Training.} 
    Domain Adversarial Training shares a training scheme similar to our method where the classifier and discriminator compete with each other~\cite{con_domain1,con_domain2}. However, its objective is to reduce the gap between the source and target distributions in the latent space. The discriminator is used to measure the divergence between these two distributions in the latent space. The training scheme of our method is also based on competition between the classifier and discriminator. Different from the previous framework, the discriminator of our method is used to capture the information of distributions of adversarial examples and clean counterparts in the latent space which helps generate the adversarial perturbations. 
    
 \textbf{GAN-based Adversarial Training Methods.}
    Several GAN-based methods leverage GANs to learn the clean data distribution and purify the adversarial examples by projecting them on the clean data manifold before classification~\cite{magnet,detect_adnet}. The framework of GAN can also be used to generate adversarial examples~\cite{gan_adv1}. The generator produces the adversarial examples to deceive both the discriminator and classifier;  the discriminator and classifier attempt to differentiate the adversaries from clean data and produce the correct labels respectively. Some adversary detector networks are proposed to detect the adversarial examples which can be well aligned with our method~\cite{detect_1,detect_2}. In these works, a pre-trained network is augmented with a binary detector network. The training of the pre-trained network and detector involves generating adversarial examples to maximize their losses. Differently, our method generates the LMAEs to minimize the loss of the discriminator and feed them as inputs to the classifier. Such adversarial examples are deemed to induce most different latent representations from the clean counterpart. 

\section{Background}

\subsection{Adversarial Training}\label{sec:at}

Adversarial training is increasingly adopted
for defending against adversarial attacks. Specifically, it solves the following minimax optimization problem through training. 
\begin{eqnarray}
\begin{aligned}
\min_{\theta}\{\mathbb{E}_{(x,y)\sim \mathcal{D}}[\max_{x'\in S_x}L(x',y;\theta)]\},
\label{eq:1}
\end{aligned}
\end{eqnarray}
where $x \in \mathbb{R}^n$ and $y \in \mathbb{R}$ are respectively the clean data samples and the corresponding labels drawn from the dataset $\mathcal{D}$, and $L(\cdot)$ is the loss function of the DNN with the model parameter $\theta \in \mathbb{R}^m$. Furthermore, we denote the clean data distribution as $Q_0$, i.e. $x\sim Q_0$., and denote $x' \in \mathbb{R}^n$ as perturbed samples in a feasible region $S_x \triangleq \{z:z\in B(x,\epsilon)\cap [-1.0, 1.0]^n\}$ with $B(z,\epsilon) \triangleq \{z: \lVert{x-z}\rVert_{\infty} \le \epsilon\}$ being the $\ell_\infty$-ball at center $x$ with radius $\epsilon$. 
By defining  $f_\theta(\cdot)$ as the mapping function from the input layer to the last latent layer, we can also rewrite the loss function of the DNN as $l(f_\theta(x), y)$ where $l(\cdot)$ denotes the loss  function calculated from the last hidden layer of the DNN, e.g. the cross entropy loss as typically used in DNN.

Whilst the outer minimization can be conducted by training to find the optimal model parameters $\theta$,
the inner maximization essentially generates the strongest adversarial attacks on a given set of model parameters $\theta$. 
In general, the solution to the minimax problem can be found by training a network minimizing the loss for worst-case adversarial examples, so as to attain adversarial robustness. 
Given a set of model parameters $\theta$, the commonly adopted solution to the inner maximization problem can lead to either one-step (e.g., FGSM) or multi-step (e.g., PGD) approach~\cite{PGD}. 
In particular, for a given single point $x$, the strongest adversarial example $x'$ at the $t$-th iteration can be iteratively obtained by the following updating rule:
\begin{eqnarray}
\begin{aligned}
x^{t+1}=\Pi_{S_x}(x^t+\alpha \cdot \mathrm{sgn}(\nabla_x L(x^t, y; \theta))),
\label{eq:2}
\end{aligned}
\end{eqnarray}
 where $\Pi_{S_x}(\cdot)$ is a projection operator to project the inputs onto the region $S_x$, $\mathrm{sgn}(\cdot)$ is the sign function, and $\alpha$ is the updating step size. For the initialization, $x^0$ can be generated by randomly sampling in $B(x,\epsilon)$. 

It appears in (\ref{eq:1}) that each perturbed sample $x'$ is obtained individually by leveraging its loss function $L(x', y; \theta)$ with its label $y$. 
However, without considering the inter-relationship between samples, we may lose the global knowledge of the data manifold structure which proves highly useful for attaining better generalization. 
This issue has been studied in a recent work \cite{feature_scattering} where a new method named feature scattering made a first step to consider the inter-sample relationship {\em within the batch}; unfortunately this approach did not take the full advantages of the global knowledge of the entire data distribution. 
In addition, relying on the maximization of the loss function, the adversarially-perturbed data samples may be highly biased toward the decision boundary, which potentially corrupts the structure of the original data distribution, especially when the decision boundary is non-optimal (see Figure~\ref{fig:toy1} again for the illustration). 

\subsection{Divergence Estimation}\label{sec:de}
To measure the discrepancy of two distributions, statistical divergence measures (e.g., Kullback-Leibler and Jensen-Shannon divergence) have been proposed. In general, given two distributions $\mathbb{P}$ and $\mathbb{Q}$ with a continuous density function $p(x)$ and $q(x)$ respectively, $f$-divergence is defined as 
$
D_f(\mathbb{P}||\mathbb{Q}) \triangleq \int_{\mathcal{X}}q(x)f\left( \frac{p(x)}{q(x)}\right)dx.
$
The exact computation of $f$-divergence is challenging, and the estimation from samples has attracted much interest. For instance, leveraging the variational methods, \cite{gen_div} proposes a method for estimating $f$-divergence from only samples; \cite{fgan} extends this method by estimating the divergence by learning the parameters of the discriminator. Specifically, the $f$-divergence between two distributions $\mathbb{P}$ and $\mathbb{Q}$ can be lower-bounded using Fenchel conjugate and Jensen's inequality \cite{fgan}. 
\begin{eqnarray}
\begin{small}
\begin{aligned}
 D_f(\mathbb{P}||\mathbb{Q})&=\int_{\mathcal{X}}q(x)\sup_{t\in \mathrm{dom} f^*} \{t \frac{p(x)}{q(x)}-f^*(t)\}dx
\\ &\geq \sup_{T\in \tau}(\int_{\mathcal{X}}p(x)T(x)dx\quad -\int_{\mathcal{X}}q(x)f^*(T(x))dx)
\\ &=\sup_W(\mathbb{E}_{x\sim \mathbb{P}}[g_f(V_W(x))] \quad +\mathbb{E}_{x\sim \mathbb{Q}}[-f^*(g_f(V_W(x)))]),
\label{f_d}
\end{aligned}
\end{small}
\end{eqnarray}
where $V_W:\mathcal{X}\to \mathbb{R}$ is a discriminator network with parameter $W$ and $g_f:\mathbb{R}\to \mathrm{dom} f^*$ is an output activation function which is determined by the type of discriminator. 
$\tau$ is an arbitrary class of functions $T:\mathcal{X}\to \mathbb{R}$. $f$ is a convex lower-semicontinuous function and $f^*$ is its conjugate defined by $f^*(t)=\sup_{u\in \mathrm{dom} f}[ut-f(u)]$. The objective of discriminator for GANs is a special case of~(\ref{f_d}) with the activation function $g_f(t)=-\log(1+e^{-t})$ and $f^*(g)=-\log(2-e^g)$. It approximates the Jensen-Shannon divergence between real and fake distributions.~\cite{wgan} also develops a method to estimate the Wasserstein distance by the neural network. In this paper, these methods will be used to estimate the Jensen-Shannon divergence between latent distributions induced by adversarial and clean examples.

\section{Adversarial Training with Latent Distribution (ATLD)}

For clarity, we first list the major notations that are used in our model.
 \begin{itemize}
        \item $X_{org}=\{x:x\sim Q_0\}$: the set of clean data samples, where $Q_0$ is its underlying distribution;
        \item $X_{p}=\{x': x'\in B(x,\epsilon), \;\forall x\sim Q_0\}$: the set of perturbed samples,  the element $x'\in X_{p}$ is in the $\epsilon$-neighborhood of the clean example $x\sim Q_0$;
        \item $f_\theta$: the mapping function from input to the latent features of the last hidden layer (i.e., the layer before the softmax layer);
        \item $Q_\theta$: the underlying distribution of the latent feature $f_\theta (x)$ for all $x \in X_{org}$;
        \item $P_\theta$: the underlying distribution of the latent feature $f_\theta (x')$ for all $x'\in X_{p}$;
        \item $\mathcal{P}$: the feasible region of the latent distribution $P_{\theta}$, which is defined as $\mathcal{P} \triangleq \{P:f_\theta(x')\sim P \text{ subject to } \forall x\sim Q_0, x'\in B(x,\epsilon)\}$. 
        \item $X_{adv}$: the set of the worst perturbed samples or Latent Manifold Adversarial Examples (LMAEs), the element $x^{adv}\in X_{adv}$ are in the $\epsilon$-neighborhood of clean example $x\sim Q_0$;
        \item $P_\theta^*$: the worst latent distribution within the feasible region $\mathcal{P}$ which leads to the largest divergence or the underlying distribution of the latent feature $f_\theta(x^{adv})$ for all $x^{adv} \in X_{adv}$;

\end{itemize}

\begin{figure*}[ht]
	\centering
	\includegraphics[width=0.8\textwidth]{{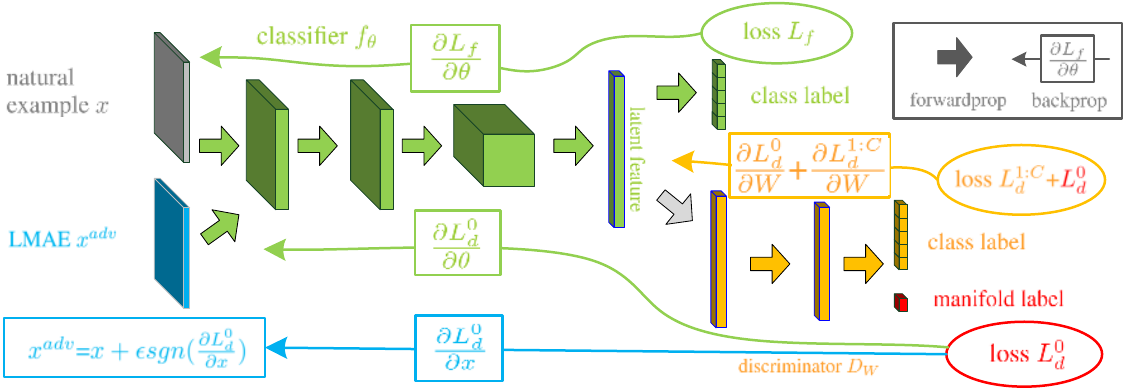}}
	\caption{Overall architecture of ATLD and its  training procedure. 1) The natural example is fed into the network, and the discriminator outputs its prediction. The manifold loss $L^0_d$ is computed to generate the LMAE $x^{adv}$ (blue arrow). 2) Both the clean and LMAE are fed into the network to train the classifier (green arrow) and the discriminator (yellow arrow) iteratively.} 
	\label{fig:framework}
\end{figure*}
As discussed in Section~\ref{sec:at}, the conventional adversarial training methods rely on the knowledge of data labels. As a result,  the local information to generate adversarial examples may be biased toward the decision boundary; such individual adversarial example generation does not capture the global knowledge of the data manifold.

To alleviate these limitations, we propose a novel notion named Latent Manifold Adversarial Examples (LMAEs) to compute the perturbed samples by leveraging the global knowledge of the entire data distribution and then disentangling them from the data labels and the loss function. Generally speaking, the perturbations are generated to enlarge the variance between latent distributions induced by clean and adversarial data. 
Formally, we try to identify the set of LMAEs $X_{adv}$ that yield in the latent space a distribution $P_\theta^*$ through $f_\theta(\cdot)$ that is the most different from the latent distribution $Q_\theta$ induced by the clean samples $X_{org}=\{x: x\sim Q_0\}$, without resorting to the corresponding labels $Y$. Again, \textit{the LMAEs are perturbed in input space but aim to `deceive' the latent manifold rather than fool the classifier as defined in the traditional adversarial examples.} It is noted that the latent space to be perturbed could be any hidden layer though it is defined in the last hidden layer before the softmax of a DNN in this paper. The optimization problem of the proposed adversarial training can then be reformulated as follows:
\begin{align}
    \min_\theta & \quad \{\mathbb{E}_{f_\theta(x^{adv})\sim P_\theta^*} [l(f_\theta(x^{adv}),y)] + D_f(P_\theta^*||Q_\theta)\} \label{eq:main-obj}\\
    \text{s.t.} & \quad P^*_\theta=\arg \max_{P_\theta \in \mathcal{P}} [D_f(P_\theta||Q_\theta)]
    \label{eq:main-constraint}
\end{align}
where  $l(\cdot)$ and $y$ are similarly defined as before, and $D_f(\cdot)$ is the $f$-divergence measure of two distributions. $\mathcal{P}=\{P:f_\theta(x')\sim P \quad \mbox{subject to} \quad \forall x\sim Q_0, x'\in B(x,\epsilon)\}$ is the feasible region for the latent distribution $P_\theta$ which is induced by the set of perturbed examples $X_p$ through $f_\theta(\cdot)$. $f_\theta(x')$ and $f_\theta(x^{adv})$ represents the latent features of the perturbed example $x'$ and the LMAE $x^{adv}$ respectively. Intuitively, we try to obtain the worst latent distribution $P_\theta^*$ which is induced by $X_{adv}$ through $f_\theta(\cdot)$ within the region $\mathcal{P}$, while the model parameter $\theta$ is learned to minimize the classification loss on the latent feature $f_\theta(x^{adv})\sim P_\theta^*$ (or equivalently LMAE $x^{adv}\in X_{adv}$) and the $f$-divergence between the latent distributions $P^*_\theta$ and $Q_\theta$ induced by LMAEs $X_{adv}$ and clean data $X_{org}$.

It is still challenging to solve the above optimization problem since both the objective function and the constraint are entangled with the LMAEs $X_{adv}$ and the model parameters $\theta$. To make the problem more tractable, we propose a novel Adversarial Training with Latent Distribution (ATLD) method. In the next subsection, by taking into account the entire data distribution globally, we first focus on the constraint and identify the LMEAs $X_{adv}$ through the maximization problem. We then solve the minimization of the objective function with the adversarial training procedure.

\subsection{Generating Latent Manifold Adversarial Examples (LMAEs) for Training}
First, we optimize the constraint Equation.~(\ref{eq:main-constraint}) to generate the adversarial examples or its induced distribution $P^*_{\theta}$ for training. 
Intuitively, the LMAEs $X_{adv}$ are crafted to maximize the divergence between the latent distributions induced by natural examples $X_{org}$ and adversarial counterparts, in which way the LMAEs would not be biased toward the decision boundary since no knowledge of labels $Y$ is required. Together with the objective function in Equation.~(\ref{eq:main-obj}), our proposed adversarial training method is to minimize such divergence as well as the classification error for LMAEs $X_{adv}$.

However, it is a challenging task to evaluate the divergence between two latent distributions. To make it more tractable, we leverage \textit{a discriminator} network for estimating the Jensen-Shannon divergence between two distributions $P_\theta^*/P_\theta$ and $Q_\theta$ according to Section~\ref{sec:de}.  
It is noted again that the class label information is not used for generating adversarial examples. Hence the adversarial examples are still unsupervised and not generated toward the decision boundary.
Then, by using Equation.~(\ref{f_d}), the optimization problem in Equation.~(\ref{eq:main-obj}) and Equation.~(\ref{eq:main-constraint}) can be approximated as follows in a tractable way. 
\begin{eqnarray}
\begin{small}
\begin{aligned}
\min_\theta & \Big\{ \sum_{i=1}^N\underbrace{L(x^{adv}_i,y_i;\theta)}_{L_f}+\\
&\sup_W\sum_{i=1}^N[\underbrace{\log D_W(f_{\theta}(x^{adv}_i))+(1-\log D_W(f_{\theta}(x_i)))}_{L_d}]\Big\} 
\\
\text{s.t.} & \quad x_i^{adv}=\arg \max_{x'_i\in B(x_i,\epsilon)}[\log D_W(f_{\theta}(x'_i))  \\&\qquad\qquad\qquad\qquad\qquad +(1-\log D_W(f_{\theta}(x_i)))] 
\label{re_p1}
\end{aligned}
\end{small}
\end{eqnarray}
where $N$ denotes the number of training samples and $D_W$ denotes the discriminator network with the sigmoid activation function and parameter $W$. $f_\theta(x_i)$ is the latent feature of the clean sample $x_i$. 
$D_W$ is used to determine whether the latent feature is from the adversary manifold (output the manifold label of the latent feature). For ease of description, we represent the components in Equation.~(\ref{re_p1}) as two parts: $L_f$ and $L_d$. $L_d$ is the manifold loss and $L_f$ represents the loss function of the classifier network. 

We now interpret the above optimization problem.
By comparing Equation.~(\ref{re_p1}) and Equation.~(\ref{eq:main-obj}), it is observed that  the Jensen-Shannon divergence between $P_\theta^*$ and $Q_\theta$ is approximated by  $\sup_W\sum_{i=1}^NL_d$, and the minimization of  the classification loss on adversarial examples is given by $\min_{\theta}\sum_{i=1}^NL_f$. The Equation.~(\ref{re_p1}) is optimized by  updating parameters $\theta$ and $W$  and crafting LMAEs $\{x_i^{adv}\}_{i=0}^N$ iteratively. The whole training procedure can be viewed as a game among three players: the classifier, discriminator, and LMAEs. The discriminator $D_W$ is learned to differentiate the latent distributions of the perturbed examples and clean data via maximizing the loss $L_d$ while the classifier $f_\theta$ is trained to (1) enforce the invariance between these two distributions to confuse the discriminator $D_W$ by minimizing the loss $L_d$, and (2)  classify the LMAEs as accurately as possible by minimizing $L_f$. For each training iteration, the LMAEs are crafted to make different the adversarial latent distribution and the natural one by maximizing $L_d$. Although $D_W$ cannot measure the divergence between the two latent distributions exactly at the first several training steps, it can help evaluate the divergence between distributions induced by perturbed examples and clean ones when the parameters $W$ converges.

However, when the latent distributions are multi-modal, which is a real scenario due to the nature of multi-class classification, it is challenging for the discriminator to measure the divergence between such distributions. Several works reveal that there is a high risk of failure in using the discriminator networks to measure only a fraction of components underlying different distributions~\cite{gan_pri,gan_mode}. \cite{gan_eq} also shows that two different distributions are not guaranteed to be identical even if the discriminator is fully confused. To alleviate such the problem, we additionally train the discriminator $D_W$ to predict the class labels for latent features as~\cite{con_domain1,con_domain2}. As a result,  the Equation.~\ref{re_p1} can then be reformulated as: 
\begin{eqnarray}
\begin{small}
\begin{aligned}
\min_\theta & \Big\{ \sum_{i=1}^N\underbrace{L(x^{adv}_i,y_i;\theta)}_{L_f}\\&+\sup_W\sum_{i=1}^N[\underbrace{\log D^0_W(f_{\theta}(x^{adv}_i))+(1-\log D^0_W(f_{\theta}(x_i)))}_{L^0_d}]\\& +\min_{W}[\underbrace{l(D^{1:C}_W(f_{\theta}(x_i)),y_i)+l(D^{1:C}_W(f_{\theta}(x^{adv}_i)),y_i)]}_{L^{1:C}_d}\Big\} 
\\
\text{s.t.} & \quad x_i^{adv}=\arg \max_{x'_i\in B(x_i,\epsilon)}[\log D^0_W(f_{\theta}(x'_i))\\&\qquad\qquad\qquad\qquad\qquad+(1-\log D^0_W(f_{\theta}(x_i))] .
\label{re_p2}
\end{aligned}
\end{small}
\end{eqnarray}
Here $D^{0}_W$ is the first dimension of the output of the discriminator, which indicates the manifold label of the latent features; $D^{1:C}_W$ are the remaining $C$ dimensions of the output of $D_W$,  used to output the class label of the latent feature; $C$ denotes the number of classes, and
$L^0_d$ and $L^{1:C}_d$ are the manifold loss and the classification loss for the discriminator network respectively. 
(\textit{Detailed derivations for Equations.~(\ref{eq:main-obj})-(\ref{re_p1})  can be seen in the Appendix.})
The detailed training procedure of our framework is depicted in Figure~\ref{fig:framework}.

\textbf{Remarks.} It is worth noting that the labeled information is not required for generating LMAEs in an unsupervised manner. Although it is used to train the discriminator, this is not compulsory. The benefit of the label information is to better deal with multi-modal distributions. Therefore, our method prevents the perturbed examples from highly biasing toward the decision boundary and more information about the original distribution structure is preserved. In addition, since the discriminator is trained with the whole dataset (both clean and adversarial examples), it captures the global information of the data manifold. Consequently, by training with LMAEs generated according to the manifold loss of the discriminator, our method can improve the model robustness against adversarial examples with the global structure of data distribution.

\subsection{Inference with Manifold Transformation}\label{sec:IMT}

To enhance the generalization of ATLD, we further develop a new inference method with manifold transformation. Although adversarially-trained models can well recognize  adversarial examples, there are still potential examples which are easily misclassified especially for unseen data. In other words, the generalization to adversarial examples is hard to achieve due to the more complex distribution of adversarial examples~\cite{ad_generalization,ad_gen2}. 

\begin{figure}[h]
	\centering
	\subfloat[\small Clean data]{
		\includegraphics[width=0.15\textwidth]{{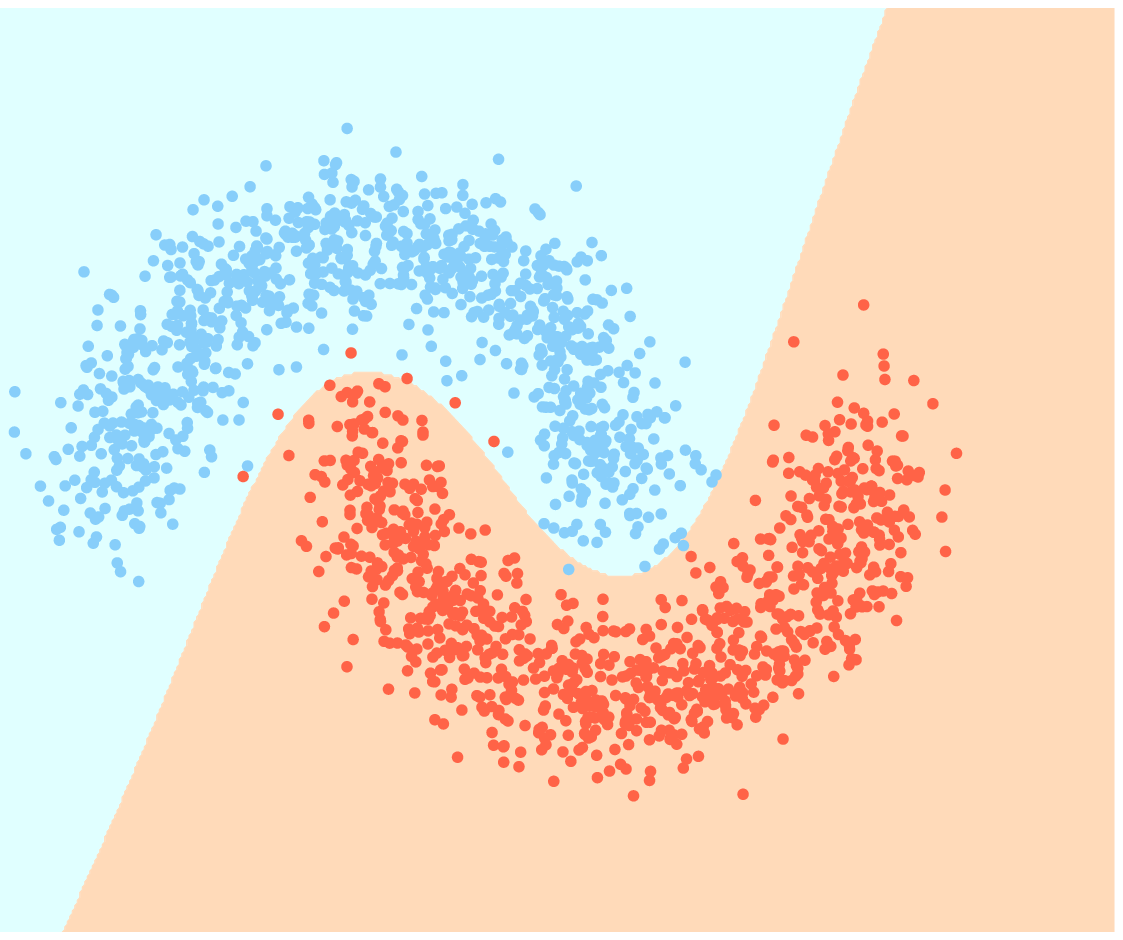}}
		\label{subfig:toy4_pgd}
	} \hspace{-5pt}
	\subfloat[\small PGD ]{
		\includegraphics[width=0.15\textwidth]{{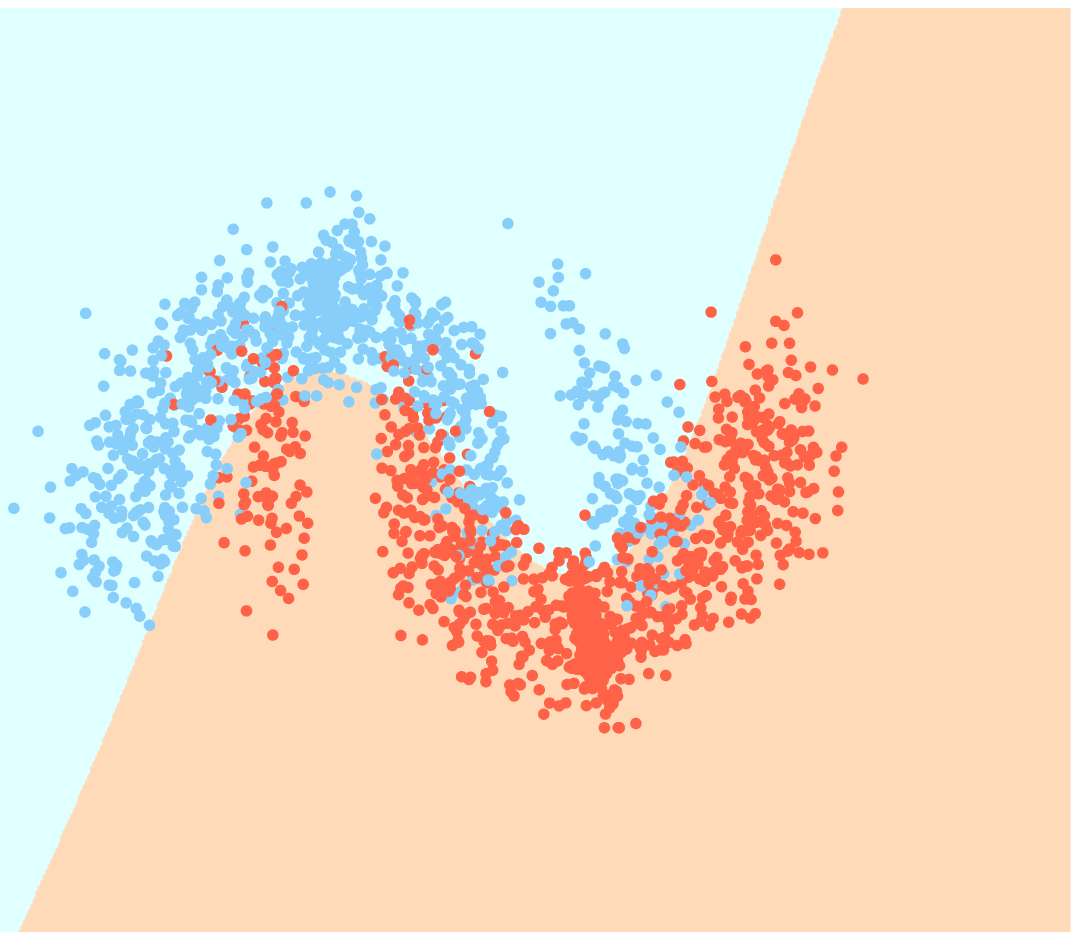}}
		\label{subfig:toy4_fs}
	} \hspace{-5pt}
	\subfloat[\small IMT]{
		\includegraphics[width=0.15\textwidth]{{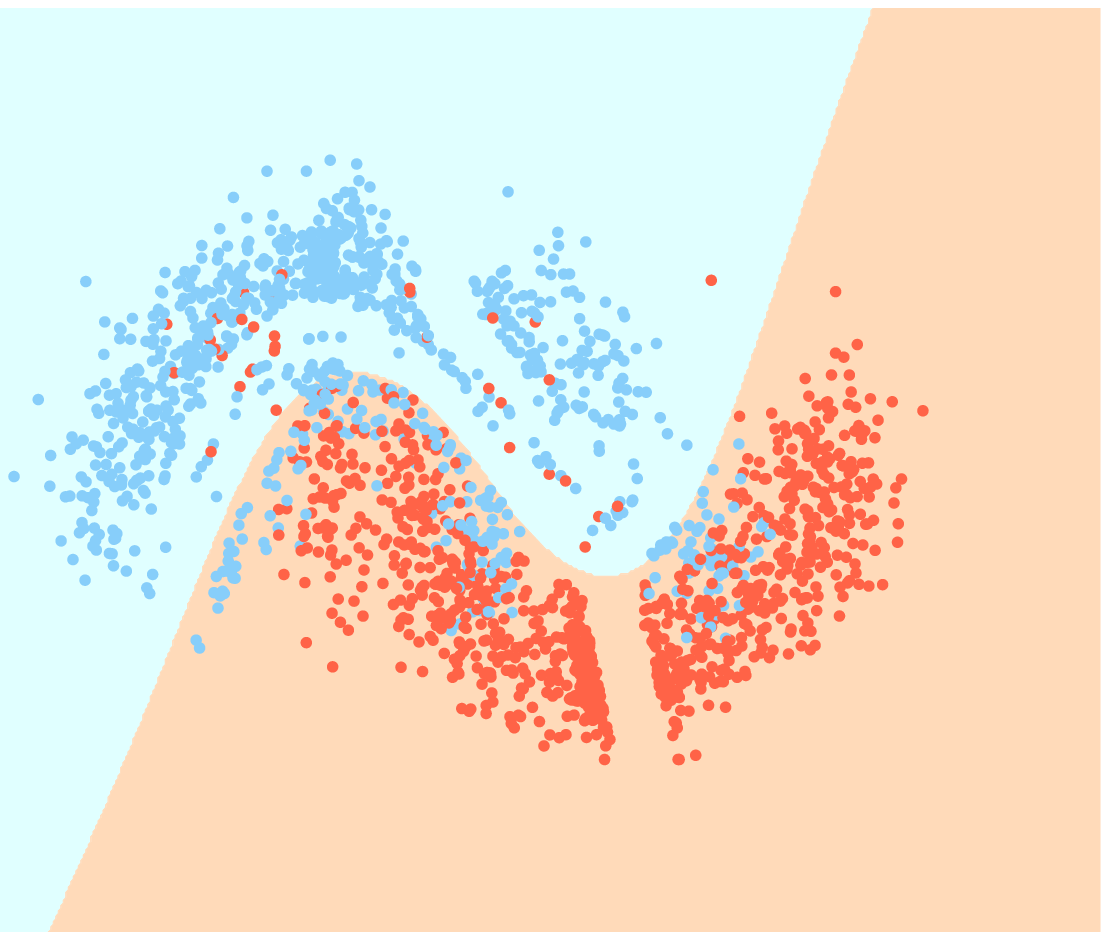}}
		\label{subfig:toy4_ours}
	}
	
	\caption{Illustration of IMT. The decision boundary is given by ATLD in all the three sub-figures, with (a) clean data, (b) perturbed data attacked by PGD, and (c) adjusted data by IMT.} 
	\label{fig:toy4}
	\vspace{-5pt}
\end{figure}

To alleviate this problem, our proposed inference method first pushes  adversarial examples toward the manifold of natural examples which is simpler and further away from the decision boundary than the adversarial distribution. Then the more separable adjusted examples  are classified by the adversarially-trained model. Specifically,  the input sample is fed into our adversarially-trained model and the discriminator outputs the probability of such a sample lying on the adversarial manifold. If this probability is higher than a certain threshold, we compute the transformed example $x^t$ by adding the specific perturbation $r^*$ to the input sample $x$ to reduce such a probability. This perturbation can be computed as: 
\begin{align}
r^*=\arg\min_{\|r\|_\infty\leq \epsilon}\log D^0_W(f_\theta(x+r)).
\label{per}
\end{align}
Intuitively, the reduction of probability of this data point lying on the adversarial manifold indicates that this point moves toward the benign example manifold after adding perturbation $r^*$. In other words, it becomes more separable since the benign example manifold is further away from the decision boundary as shown in Figure~\ref{fig:toy4}. When the probability of the image lying on the adversary manifold is lower than the threshold, we still add such a perturbation to the input image to make it more separable but with a smaller magnitude. In the experiments,  we show this perturbation can  move the adversarial examples away from the decision boundary.

\section{Experiments}

We conduct experiments on the widely-used 
datasets, CIFAR-10, SVHN, and CIFAR-100.
Following the Feature Scattering~\cite{feature_scattering}, we leverage the wideresnet~\cite{wideresnet} as our basic classifier and discriminator model structure. The initial learning rate is empirically set to $0.1$ for all three datasets. We train our model $400$ epochs on Pytorch and RTX2080TI with transition epoch ${60,90}$ and decay rate $0.1$. The input perturbation budget is set to $\epsilon = 8$ with the label smoothing rate as $0.5$. We use $L_\infty$ perturbation in this paper including all the training and evaluation.

We evaluate the various models on white-box and black-box attacks and report robust accuracy of our proposed models: ATLD (with IMT) and ATLD$+$ (with limited IMT). Under the white-box attacks, we compare the accuracy of the proposed method with several competitive methods, including (1) the original wideresnet (Standard) trained with natural examples;
(2) Traditional Adversarial Training with PGD (AT)~\cite{PGD}; (3) Triplet Loss Adversarial training (TLA)~\cite{metric_AT}; (4) Layer-wise Adversarial Training (LAT): injecting adversarial perturbation into the latent space~\cite{harnessing}; (5) Bilateral:
adversarial perturb on examples and labels both~\cite{Bilateral}; (6) Feature-Scattering (FS): generating adversarial examples with considering inter-relationship of samples~\cite{feature_scattering}. These algorithms present the most competitive performance in defending against adversarial attacks. Under the black-box attacks, we compare four different algorithms used to generate the test time attacks: Vanilla training with natural examples, adversarial training with PGD, FS, and our proposed model.

\subsection{Ablation Study}
\begin{figure*}[htbp]
	\centering
	\subfloat[\small CIFAR-10]{
		\includegraphics[width=0.3\textwidth]{{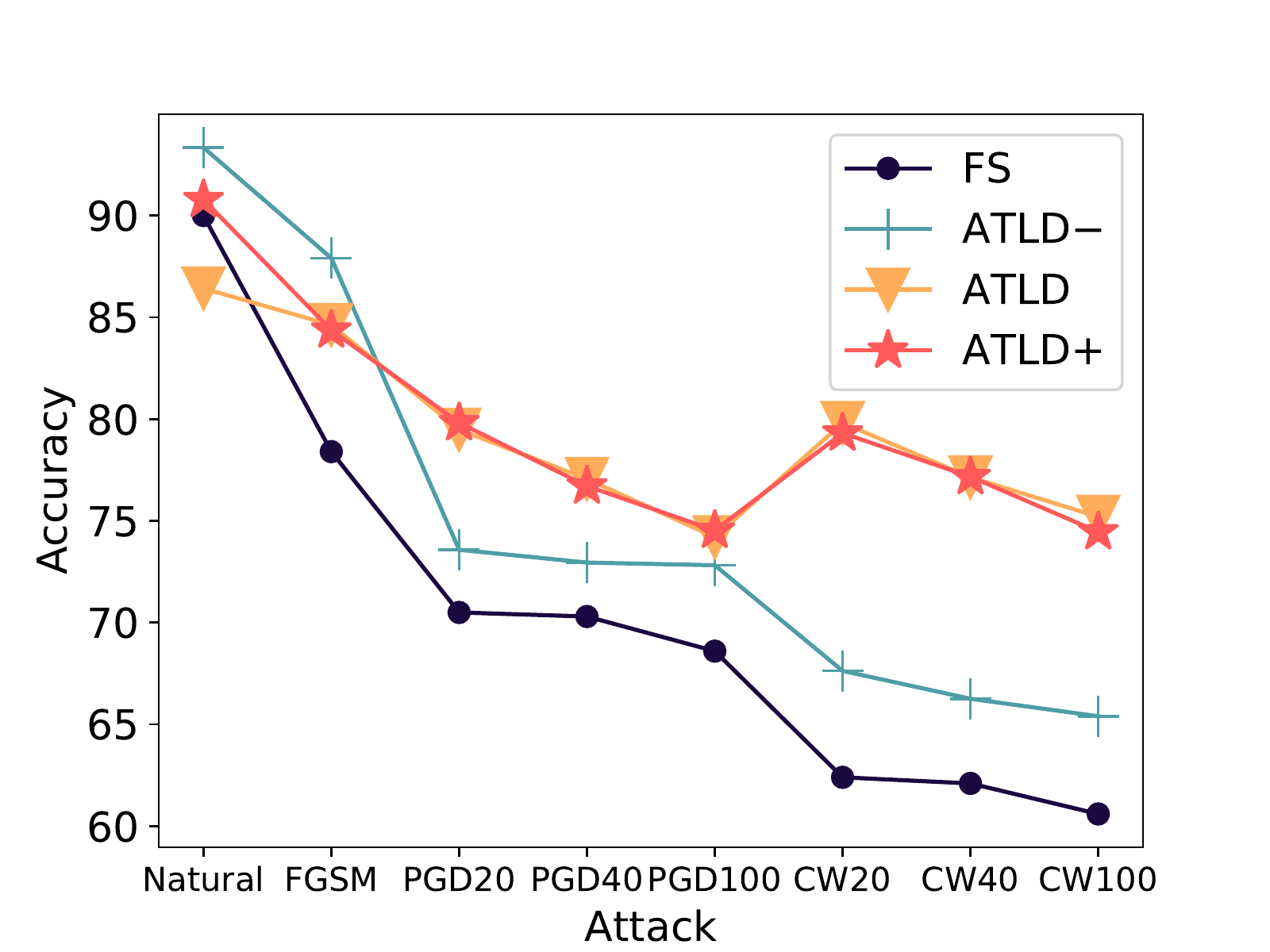}}
		\label{subfig:ab_cifar10}
	} 
	\subfloat[\small CIFAR-100 ]{
		\includegraphics[width=0.3\textwidth]{{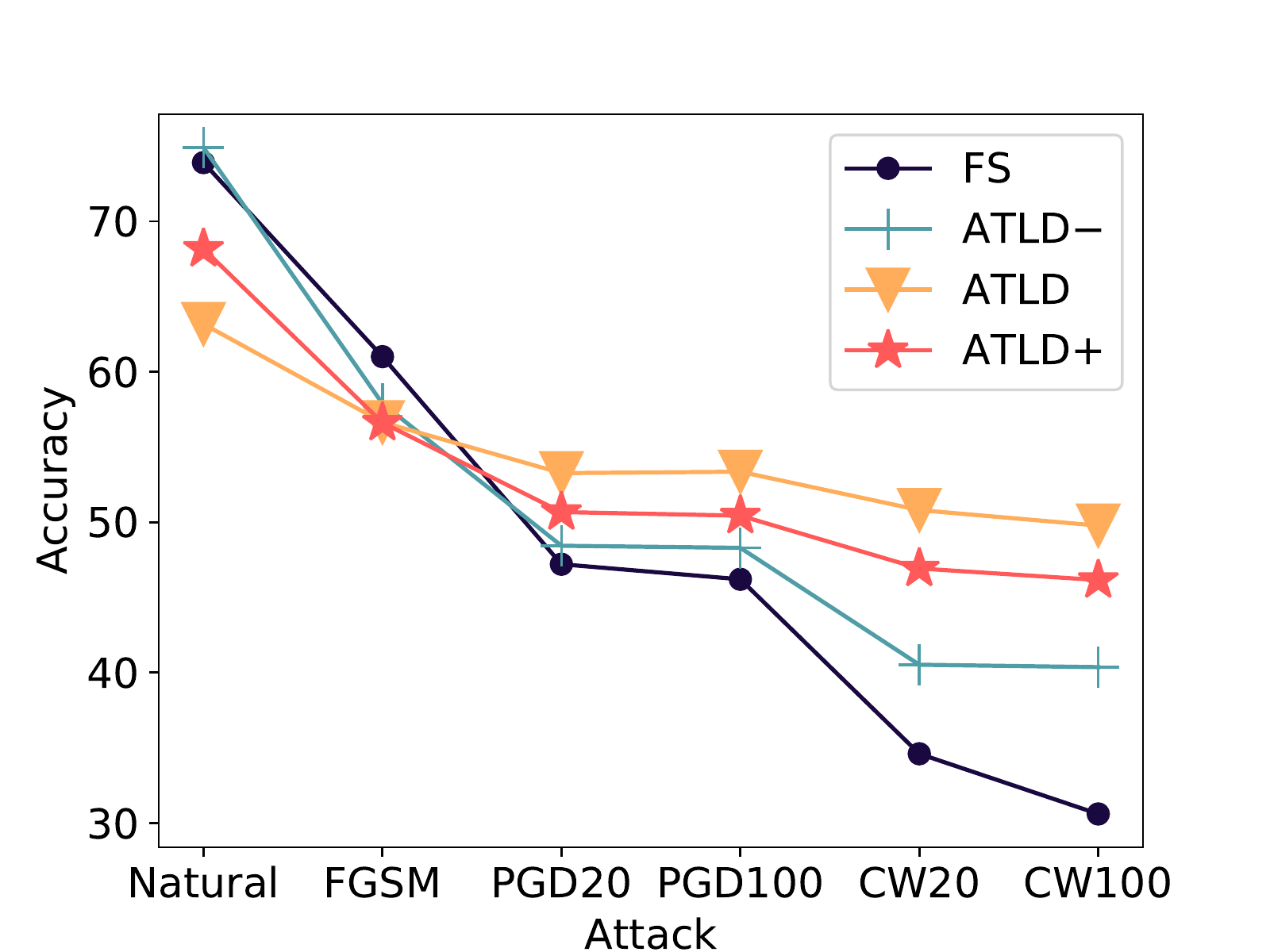}}
		\label{subfig:ab_cifar100}
	} 
	\subfloat[\small SVHN]{
		\includegraphics[width=0.3\textwidth]{{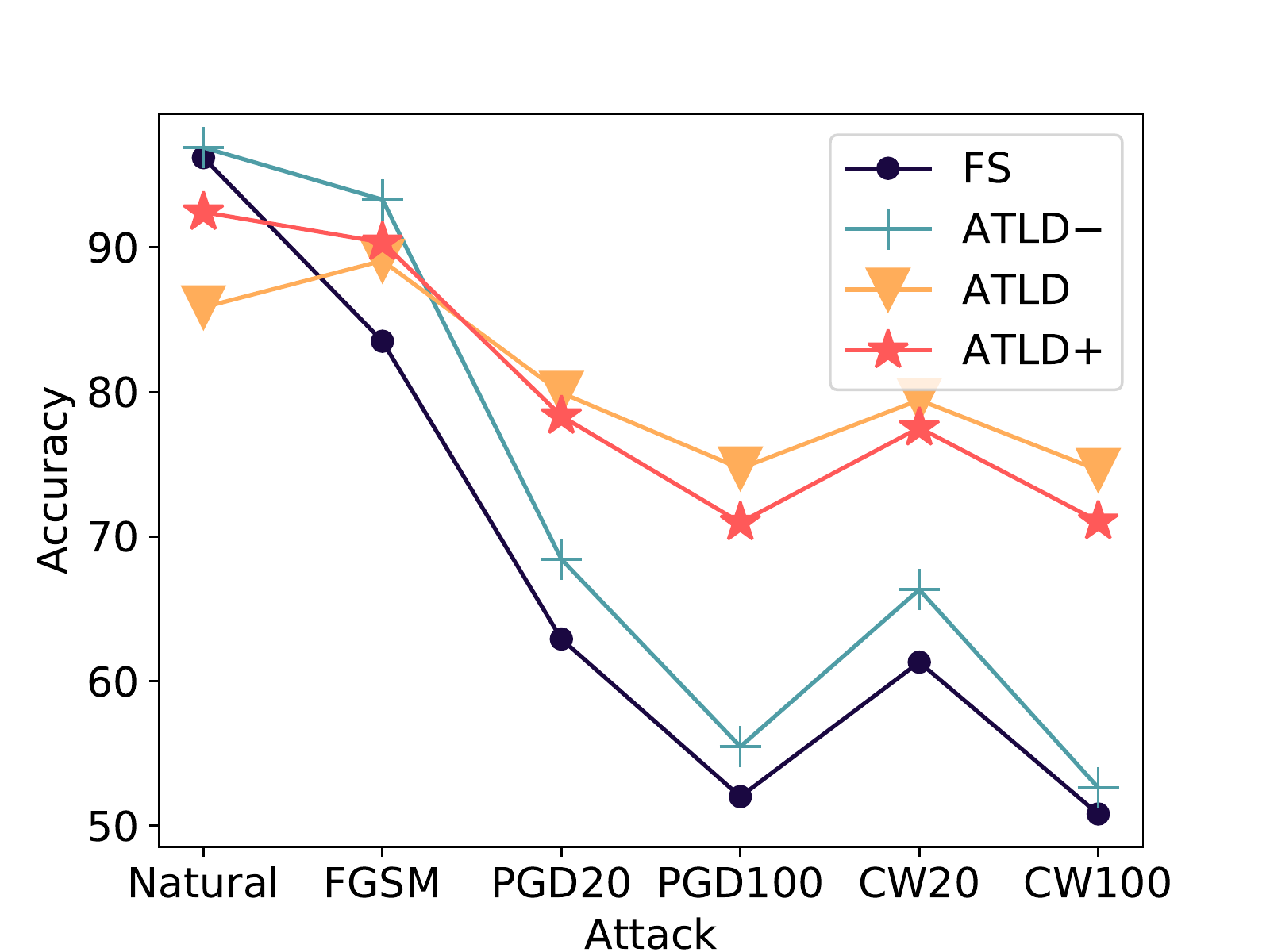}}
		\label{subfig:ab_svhn}
	}
	\caption{Ablation study On CIFAR-10 and CIFAR-100} 
	\label{fig:ab1}
\end{figure*}
To better demonstrate the effectiveness and properties of ATLD and IMT, we first perform ablation experiments on defending against white-box attacks with $\epsilon=8$ on CIFAR-10, CIFAR-100, and SVHN as shown in Figure~\ref{fig:ab1}.  As observed, ALTD$-$ (the ALTD model without applying IMT) outperforms FS in general. It deserves our attention that considering the manifold information, our proposed ATLD$-$ has better performance on clean data on all three datasets. However, IMT appears to have a negative impact on natural and FGSM data while could significantly boost the performance against PGD and CW. In order to reduce the impact of IMT on the natural and FGSM data, a threshold is used to limit the  perturbation of IMT based on the output of the discriminator. The perturbation is halved if the output of the discriminator is within the range of $[0.3, 0.7]$ (ATLD+). Under such a setting, our approach could achieve high performance on adversarial attacks without sacrificing too much accuracy on clean data. 

In practice, our proposed ATLD would not need expensive computation compared with other methods. For example, compared with FS that needs expensive computation on the optimal transport and
AT that requires multiple backpropagations, our proposed ATLD training from scratch both for the classifier and the discriminator does not incur extra computations. According to the experiments on CIFAR-10 with GPU (NVIDIA RTX-2080TI), for each epoch, ATLD
takes about 21m37s, which is less than FS (25m14s) and standard AT with PGD-7 (28m53s). Furthermore, our proposed IMT can give real-time predictions, as the additional computational cost introduced by solving the optimization problem appears negligible, compared with data reading and transmission between CPU and GPU. In our experiments, on RTX-2080TI and CIFAR-10, the difference in inference time for the whole test dataset between ATLD (25s) and ATLD$-$(without IMT) (25s) is indistinguishable on natural data.

In this paper, we take the ATLD and ATLD$+$ as our main contributions. We will mainly focus on these two methods in the later discussion.

\subsection{Defending White-box Attacks}

\begin{table}[htbp]
\caption{Robust accuracy under different white-box attacks on CIFAR-10}
\label{cifar-table}
\begin{center}
\begin{small}
\begin{sc}
\scalebox{0.7}
{
\begin{tabular}{lcccccc}
\toprule
	\multirow{2}{*}{Models} & \multirow{2}{*}{Clean} & \multicolumn{4}{c}{Accuracy under White-box Attack ($\epsilon$ = 8)}\\
	\cmidrule(r){3-6}
	& & FGSM & ~~\,PGD$\mid$CW20 & ~~\,PGD$\mid$CW40 & ~~~~\,PGD$\mid$CW100\\
\midrule
Standard    & \textbf{95.60} &36.90 &0.00$\mid$0.00 &0.00$\mid$0.00 &0.00$\mid$0.00 \\
AT & 85.70 &54.90 &44.90$\mid$45.70 &44.80$\mid$45.60 &44.80$\mid$45.40\\
TLA &86.21 &58.88 &51.59$\mid$~~~~-\,~~~~ & ~~~~-~~~~$\mid$~~~~-~~~~ &~~~~-~~~~$\mid$~~~~-~~~~ \\
LAT &87.80 &-& 53.84$\mid$53.04 &~~~~-~~~~$\mid$~~~~-~~~~~& ~~~~-~~~~$\mid$~~~~-~~~~~\\
Bilateral & 91.20 &70.70 &57.50$\mid$56.20& ~~~~-~~~~$\mid$~~~~-~~~~ &55.20$\mid$53.80 \\
TRADES & 86.76 & 66.12 & 51.80$\mid$49.86 & 51.60$\mid$49.77 & 51.54$\mid$49.70\\
FS  & 90.00 &78.40 &70.50$\mid$62.40 &70.30$\mid$62.10& 68.60$\mid$60.60 \\
\hline
ATLD & 86.42 & \textbf{84.62} & 79.48$\mid$\textbf{79.81} & \textbf{77.06}$\mid$77.14 & 74.20$\mid$\textbf{75.20}\\
ATLD$+$ & 90.78 & 84.37 & \textbf{79.82}$\mid$79.31 & 76.71$\mid$\textbf{77.17} & \textbf{74.53}$\mid$74.46 \\

\bottomrule
\end{tabular}
}
\end{sc}
\end{small}
\end{center}
\vskip -0.1in
\end{table}

\begin{table}[t]
\caption{Robust accuracy under different white-box attacks  on CIFAR-100 and SVHN}
\label{svhn-table}
\begin{center}
\begin{small}
\begin{sc}
\scalebox{0.8}
{
\begin{tabular}{lcccccccc}
\toprule
	\multirow{2}{*}{Models}  & \multicolumn{4}{c}{CIFAR-100 ($\epsilon$ = 8)} \\
	\cmidrule(r){2-5} & CLEAN & FGSM  & ~~\,PGD$\mid$CW20 & ~~~~\,PGD$\mid$CW100 & \\
\midrule
Standard &\textbf{79.00} &10.00 &0.00$\mid$0.00 &0.00$\mid$0.00\\
AT &59.90 &28.50 &22.60$\mid$23.20 &22.30$\mid$23.00\\
LAT &60.94 &-&27.03$\mid$~~~~\,-~~~~ &26.41$\mid$~~~~\,-~~~~~\\
Bilateral &68.20 &60.80 &26.70$\mid$~~~~\,-~~~~ &25.30$\mid$22.10 \\
TRADES & 61.46 & 39.45 & 30.54$\mid$27.19 & 30.55$\mid$27.12 \\
FS &73.90 &\textbf{61.00} &47.20$\mid$34.60 &46.20$\mid$30.60\\
\hline
ATLD & 63.17 & 56.64 & \textbf{53.26}$\mid$\textbf{50.80} & \textbf{53.35}$\mid$\textbf{49.77} \\
ATLD$+$ & 68.17 & 56.61 & 50.67$\mid$46.92 & 50.43$\mid$46.15\\
\midrule
	\multirow{2}{*}{Models}  & \multicolumn{4}{c}{SVHN ($\epsilon$ = 8)}  \\
	\cmidrule(r){2-5} 
	& CLEAN & FGSM  & ~~\,PGD$\mid$CW20 & ~~~~\,PGD$\mid$CW100 \\
\midrule
Standard & \textbf{97.20} & 53.00 & 0.30$\mid$0.30 & 0.10$\mid$0.10\\
AT & 93.90 & 68.40 &47.90$\mid$48.70 &46.00$\mid$47.30\\
LAT & 91.65& - &60.23$\mid$~~~~\,-~~~~~& 59.97$\mid$~~~~\,-~~~~~\\
Bilateral & 94.10 &69.80 &53.90$\mid$~~~~\,-~~~~ &50.30$\mid$48.90 \\
TRADES & 93.90 & 77.42 & 61.54$\mid$58.05 & 60.75$\mid$57.66  \\
FS & 96.20 & 83.50 &62.90$\mid$61.30&52.00$\mid$50.80 \\
\hline
ATLD & 85.85 & 89.07 & \textbf{79.93}$\mid$\textbf{79.44} & \textbf{74.70}$\mid$\textbf{74.59}\\
ATLD$+$ &92.43 & \textbf{90.34} & 78.33$\mid$77.52 & 70.96$\mid$71.04 \\
\bottomrule
\end{tabular}
}
\end{sc}
\end{small}
\end{center}
\vskip -0.1in
\end{table}

We show the classification accuracy under several white-box attacks on CIFAR-10, CIFAR-100, SVHN in this section. We first report the accuracy on CIFAR-10 in Table~\ref{cifar-table} with the attack iterations $T=20, 40, 100$ for PGD~\cite{PGD} and CW~\cite{cw_attack} attacks.

As observed from Table~\ref{cifar-table}, overall, our proposed method achieves a clear superiority over all the defence approaches on adversarial examples and compatible accuracy on clean data on CIFAR-10.  We observe that the standard model performs the best on clean data, but our proposed ATLD$+$ achieves the second clean accuracy compared to other defence methods (marginally below to Bilateral). Our approach performs much better than the other baseline models on PGD and CW attacks. Particularly, with the implementation of Inference with Manifold Transformation (IMT), our approach ATLD is $8.9\%$ and $17.4\%$ higher than the Feature Scattering under PGD20 and CW20 attack respectively. 

The accuracy on CIFAR-100 and
SVHN are shown in Table~\ref{svhn-table} with the attack iterations $T = 20, 100$ for both PGD and CW for conciseness. Although our method is slightly weaker than Feature Scattering under FGSM attack on CIFAR-100, overall, our proposed method ATLD achieves state-of-the-art performance over all the other approaches under various adversarial attacks, specifically exceeding Feature Scattering by almost $19.2\%$ and $23.8\%$ against the attack of CW100 on CIFAR-100 and SVHN respectively. 

\subsection{Defending Black-box Attacks}

\begin{table*}[htbp]
\caption{Robust accuracy under transfer-based black-box attacks}
\label{black-table}
\begin{center}
\begin{small}
\begin{sc}
\scalebox{0.75}
{
\begin{tabular}{lcccccccccccccccccccccc}
\toprule
	\multirowcell{3}{Defense\\Models}  & \multicolumn{4}{c}{Attacked Models (CIFAR-10)} & \multicolumn{4}{c}{Attacked Models (CIFAR-100)}  & \multicolumn{4}{c}{Attacked Models (SVHN)} \\
    \cmidrule(r){2-5} \cmidrule(r){6-9} \cmidrule(r){10-13} &\multicolumn{2}{c}{Vanilla Training} & \multicolumn{2}{c}{Adversarial Training} & \multicolumn{2}{c}{Vanilla Training} & \multicolumn{2}{c}{Adversarial Training} & \multicolumn{2}{c}{Vanilla Training} & \multicolumn{2}{c}{Adversarial Training}\\
	\cmidrule(r){2-3} \cmidrule(r){4-5} \cmidrule(r){6-7} \cmidrule(r){8-9} \cmidrule(r){10-11} \cmidrule(r){12-13}
	 & FGSM  & PGD$\mid$CW20& FGSM  & PGD$\mid$CW20& FGSM  & PGD$\mid$CW20& FGSM  & PGD$\mid$CW20& FGSM  & PGD$\mid$CW20 & FGSM  & PGD$\mid$CW20\\
\midrule
AT & 83.59 & 84.40$\mid$84.47 & 77.63 & 74.28$\mid$73.30 & \textbf{59.57} & 60.30$\mid$60.24 & 56.62 & 55.42$\mid$56.58 & 88.31 & 89.54$\mid$89.60 & 77.65 & 73.43$\mid$74.34 \\

TRADES & \textbf{84.67} & 85.35$\mid$85.28 & 77.71 & 74.11$\mid$73.99 & 59.29 & 59.52$\mid$59.93 & 54.90 & 53.70$\mid$55.20 & \textbf{90.47} & 91.90$\mid$91.89 & 83.29 & 78.10$\mid$78.79\\
\hline
ATLD & 82.85 & 85.35$\mid$85.67 & 82.81 & 76.34$\mid$77.36 & 56.51 & 57.14$\mid$57.32 & 57.46 & 55.14$\mid$56.97 & 89.40 & 91.74$\mid$92.15 & 91.05 & 87.25$\mid$88.39\\

ATLD+ & 84.21 & \textbf{87.32}$\mid$\textbf{87.12} & \textbf{84.63} & \textbf{80.11}$\mid$\textbf{80.70} & 56.31 & \textbf{61.14}$\mid$\textbf{62.08} & \textbf{58.15} & \textbf{60.62}$\mid$\textbf{61.56} & 90.40 & \textbf{92.71}$\mid$\textbf{92.97} & \textbf{92.24} & \textbf{90.03}$\mid$\textbf{90.60}\\
\bottomrule
\end{tabular}
}

\end{sc}
\end{small}
\end{center}
\vskip -0.1in
\end{table*}
To further verify the robustness of ATLD, we conduct transfer-based black-box attack experiments on CIFAR-10, CIFAR-100 and SVHN. Two different agent models (Resnet-18) are used for generating test time attacks including the Vanilla Training model, and the Adversarial Training with PGD model. 
As demonstrated by the results in Table~\ref{black-table}, our proposed approach can achieve competitive performance almost in all cases. Specifically, ATLD$+$ outperforms the AT and TRADES in 15 out of 18 cases while it demonstrates comparable or slightly worse accuracy in the other 3 cases. The performance of our two methods shows marginally inferior to AT or TRADES against FGSM data, while our method outperforms AT and TRADES significantly against PGD20 and CW20 adversarial attacks both on the Vanilla Training model, and the Adversarial Training with PGD model.




\subsection{Model Robustness Against PGD and CW attacker Under Attack Budgets}
\begin{figure*}[htbp]
	\centering
	\subfloat[PGD on CIFAR-10]{
		\includegraphics[width=0.25\textwidth]{{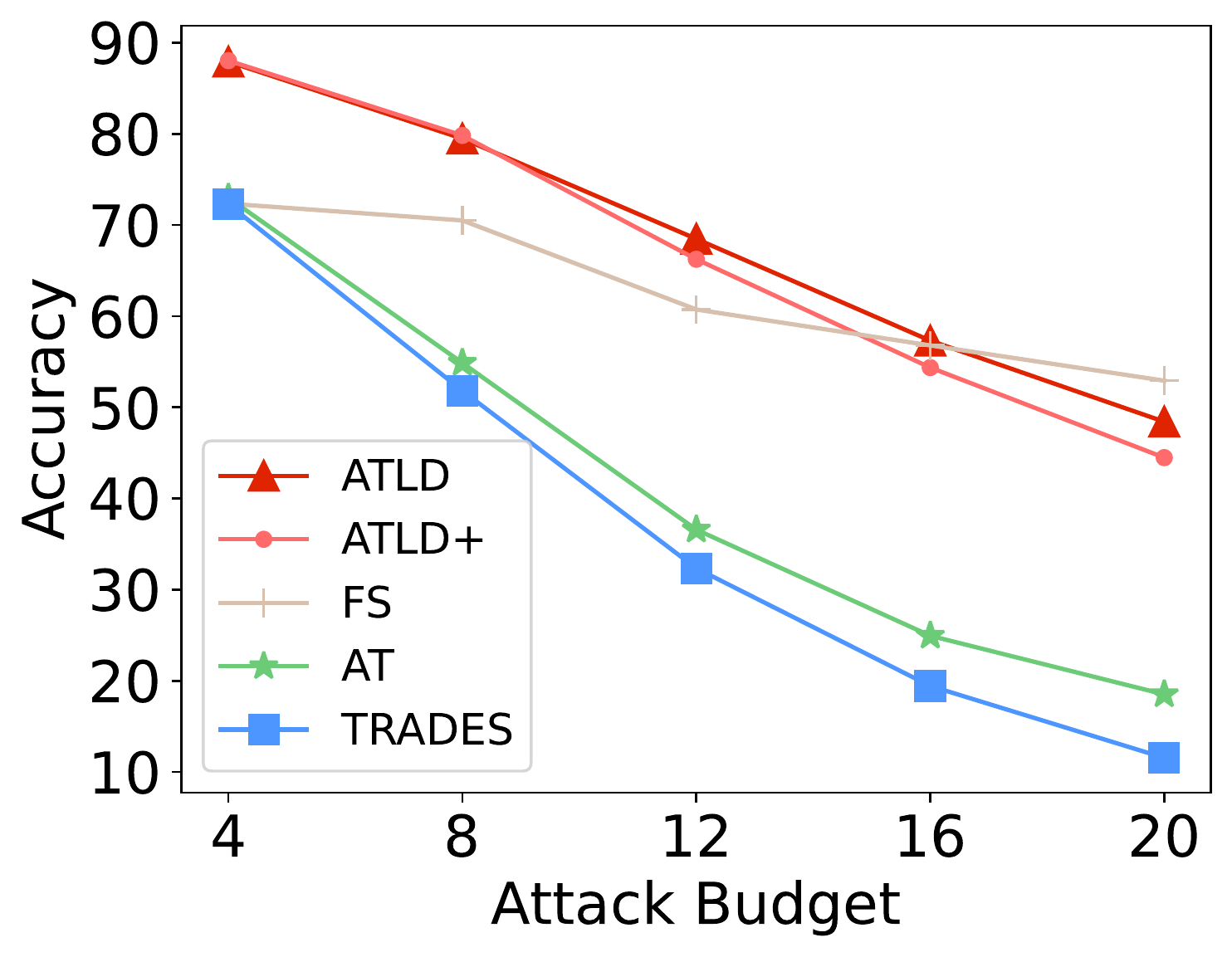}}
		\label{subfig:PGD20-cifar10}
	}
	\subfloat[PGD on CIFAR-100]{
		\includegraphics[width=0.25\textwidth]{{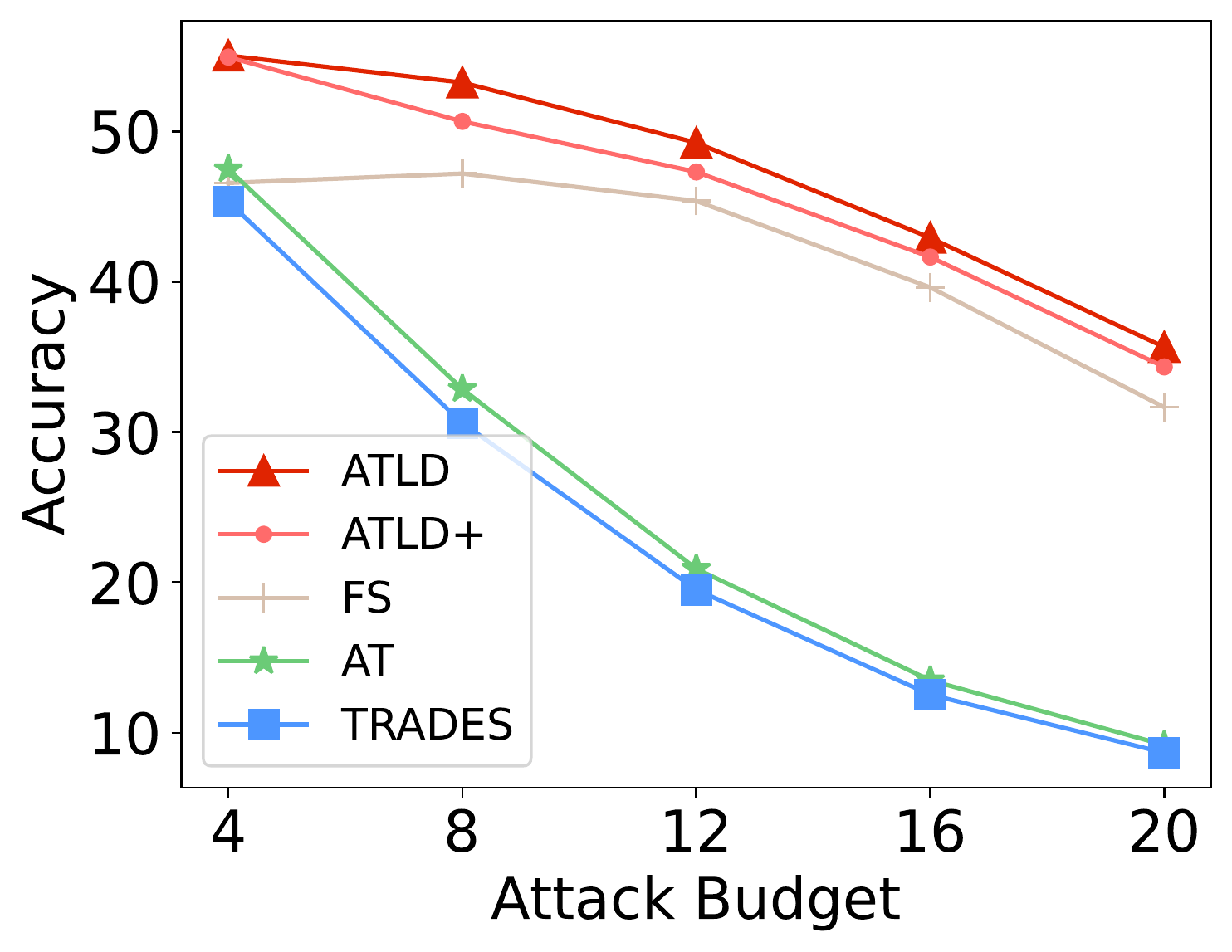}}
		\label{subfig:PGD20-cifar100}
	}
	\subfloat[PGD on SVHN]{
		\includegraphics[width=0.25\textwidth]{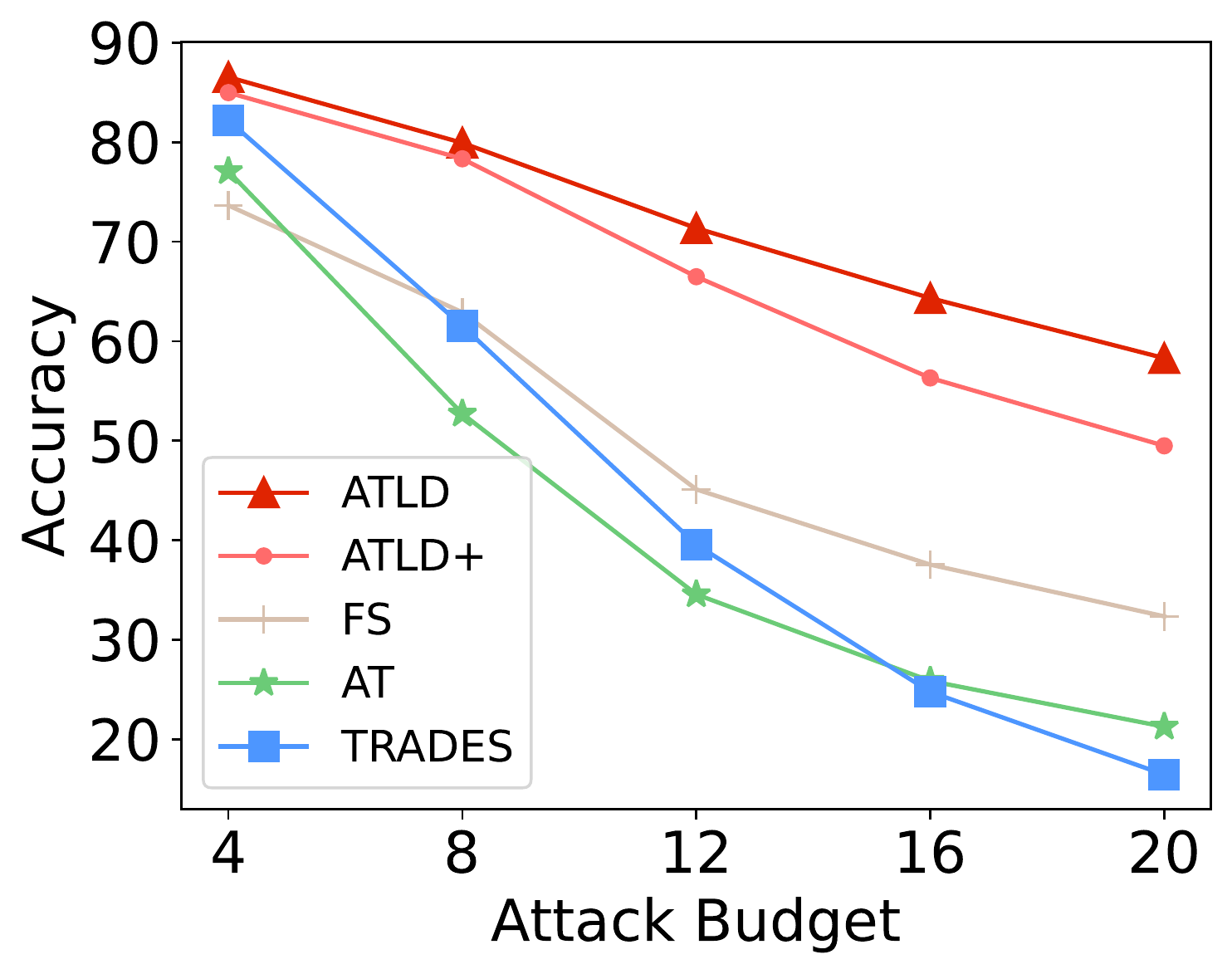}
		\label{subfig:PGD20-svhn}
	}\\
 
		\subfloat[CW on CIFAR-10]{
		\includegraphics[width=0.25\textwidth]{{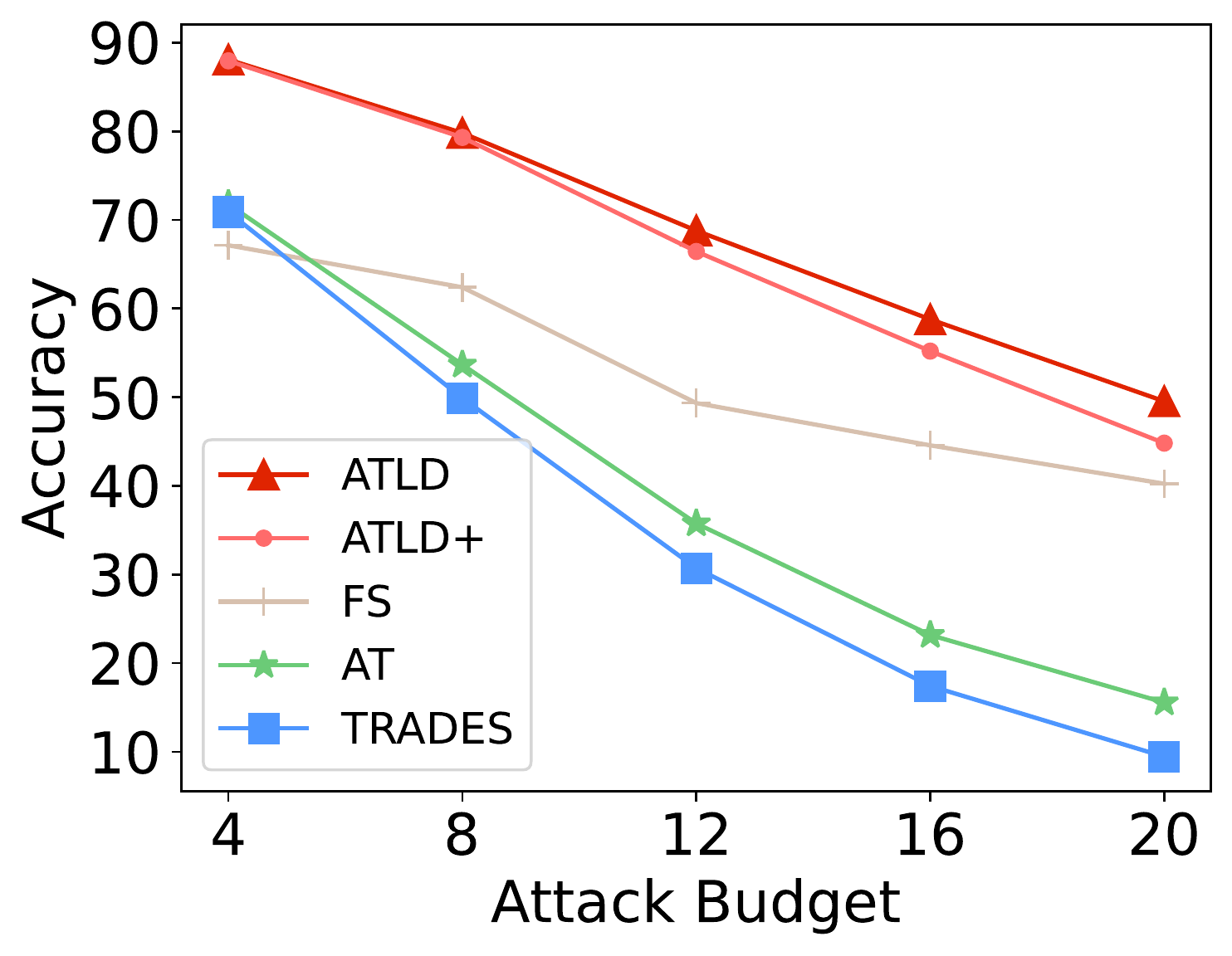}}
		\label{subfig:CW20-cifar10}
	}
	\subfloat[CW on CIFAR-100]{
		\includegraphics[width=0.25\textwidth]{{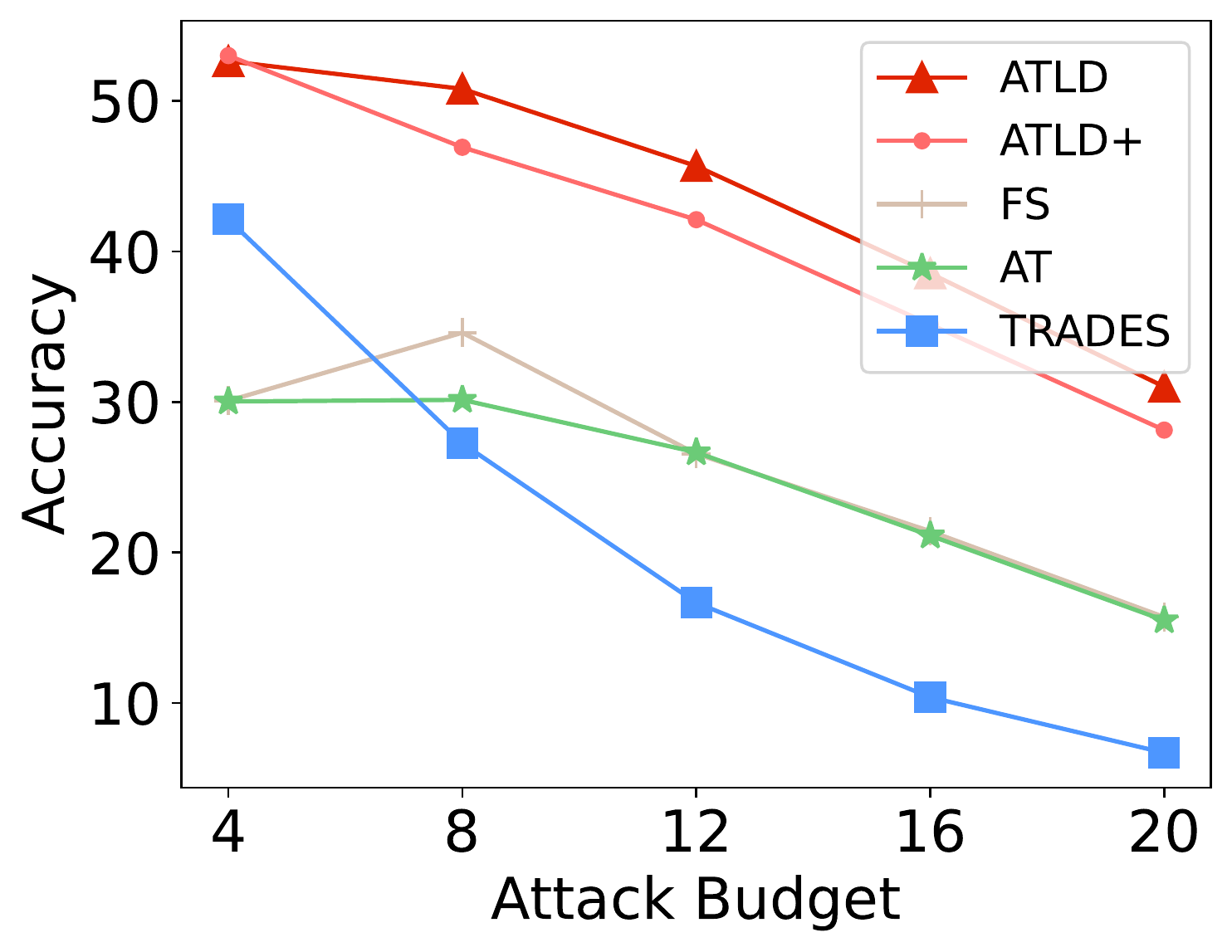}}
		\label{subfig:CW20-cifar100}
	}
	\subfloat[CW on SVHN]{
		\includegraphics[width=0.25\textwidth]{{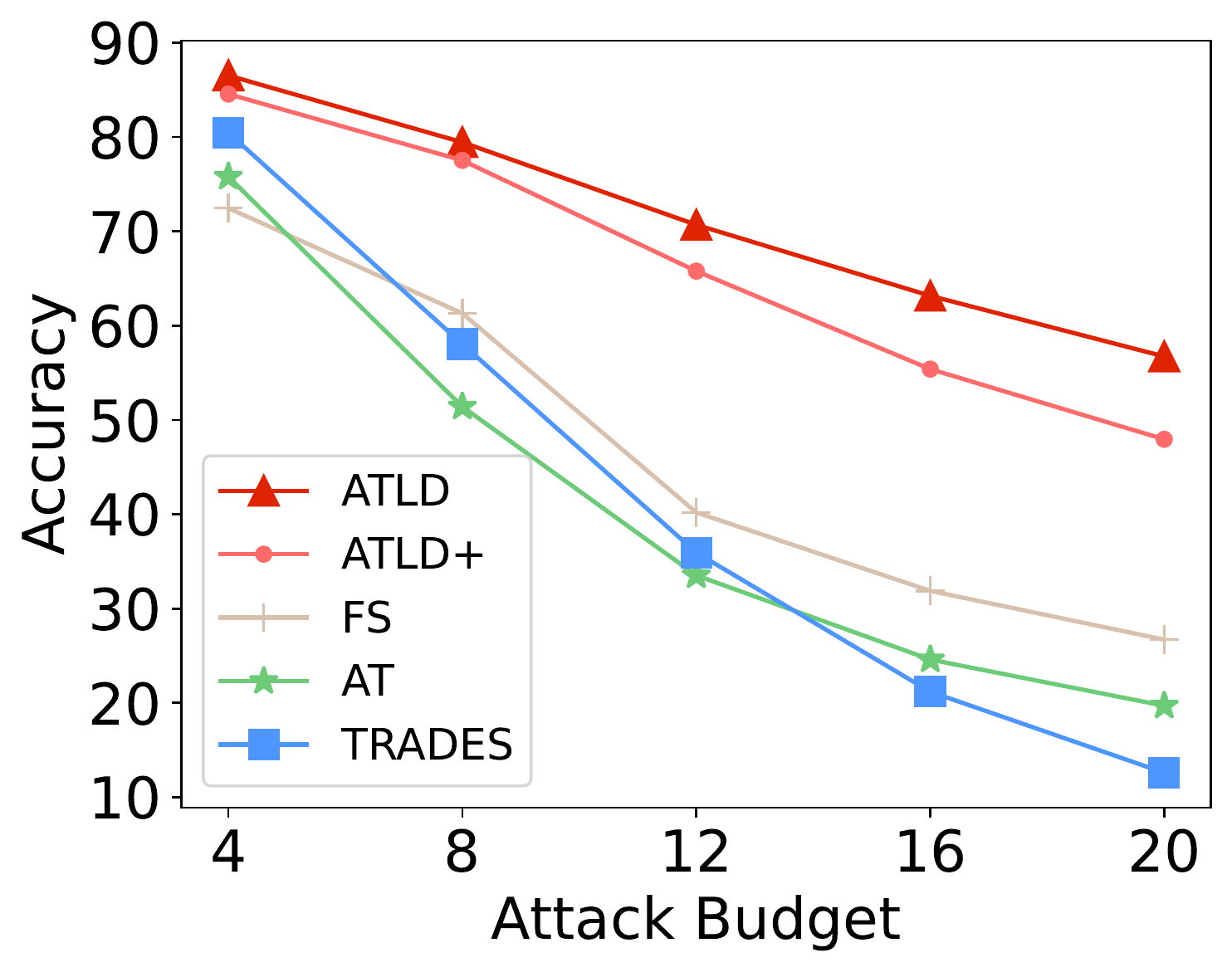}}
		\label{subfig:CW20-svhn}
	}	
	\caption{Model performance under PGD and CW attacks with different attack budgets.} 
	\label{fig:budget}
\end{figure*}
In this subsection, we examine the model robustness against PGD and CW attacks under different attack budgets with a fixed attack step of 20. These results are shown in Figure~\ref{fig:budget}. It is observed that the performance of Adversarial Training with the PGD method (AT) drops quickly as the attack budget increases. The trend and performance of TRADES are comparable to AT in most cases, except against CW20 on CIFAR-100, where TRADES substantially outperforms AT and FS when $\epsilon=4$, but is inferior to AT and FS at large $\epsilon$. The Feature Scattering method (FS) can improve the model robustness across a wide range of attack budgets. The proposed approach ATLD and ATLD$+$ further boost the performance over FS by a large margin under different attack budgets especially under CW attack, except that our ATLD and ATLD$+$ are slightly inferior to FS under PGD attack with budget $\epsilon=20$ on CIFAR-10.

\subsection{Defending Recent and Sophisticated Attacks}

\begin{table*}[htbp]
\caption{Robust accuracy under AutoAttack (AA) and RayS on CIFAR-10 and CIFAR-100}
\label{aa-table}
\begin{center}
\begin{sc}
\scalebox{0.9}
{
\begin{threeparttable}
\begin{tabular}{lcccccccccccc}
\toprule

	\multirow{2}{*}{Models}  & \multicolumn{4}{c}{CIFAR-10($\epsilon$ = 8)} & \multicolumn{4}{c}{CIFAR-100($\epsilon$ = 8)}\\
	
	\cmidrule(r){2-9} & CLEAN & CW100 & AA & RayS & CLEAN & CW100 & AA & RayS &\\
\midrule
AT & 87.14 & 50.60 & 44.04 & 50.70 & 59.90 & 23.00 & - & -\\
TRADES & 84.92 & 56.43 & 53.08 & 57.30 & - & - & - & -\\
FS & 89.98 & 60.6 & 36.64 & 44.50 & 73.90 & 30.60 & - & -\\
Robust-overfitting & 85.34 & 58.00 & 53.42 & 58.60 & 53.83 & 28.10 & 18.95 & -\\
Pretraining\tnote{*} & 87.11 & 57.40 & 54.92 & 60.10 &  59.23 & 33.50 & 28.42 & -\\
WAR\tnote{*} & 85.60 & 60.65 & 60.65 & 63.2 & - & - & - & -\\
RTS\tnote{*} & 89.69 & 62.50 & 59.53 & 64.6 & - & - & - & -\\
\cite{gowal2020uncovering} & 85.29 & 57.14 & 57.20 & - & 60.86 & 30.67 & 30.03 & -\\
\cite{gowal2020uncovering}\tnote{*} & \textbf{91.10} & 65.87 & 65.88 & -  &\textbf{69.15} & 37.70 & \textbf{36.88} & -\\
\hline
ATLD & 86.42 & \textbf{75.20} & 70.49 & 70.60 & 63.17 & \textbf{49.77} & 31.09 & 41.98 \\

ATLD+ & 90.78 & 74.46 & \textbf{70.60} & \textbf{81.68} & 68.17 & 46.15 & 32.36 & \textbf{43.91}\\

\bottomrule
\end{tabular}
 \begin{tablenotes}
        \footnotesize
        \item[*] indicates models which require additional data for training.  
      \end{tablenotes}
    \end{threeparttable}}
\end{sc}
\end{center}
\vskip -0.1in
\end{table*}

 As shown in ~\cite{autoattack,rays}, several models (such as FS) could achieve high enough robustness against PGD and CW attacks, but they may fail to defend against stronger attacks. To further evaluate the model robustness against stronger attacks, we evaluate the robustness of our proposed method ATLD and ATLD+ against AutoAttack (AA)~\cite{autoattack} and RayS~\cite{rays} attacks with $L_\infty$ budget $\epsilon=8$ on CIFAR-10 and CIFAR-100 as shown in Table~\ref{aa-table}. We compare the proposed ATLD with other the-state-of-art defence methods against AA and Rays including (7)TRADES: trading adversarial robustness off against accuracy~\cite{trades}; (8) Robust-overfitting: improving models adversarial robustness by simply using early stop~\cite{overfitting}; (9) Pretraining: improving models adversarial robustness with pre-training~\cite{pretraining}; (10)WAR: mitigating the perturbation stability deterioration on wider models~\cite{wu2020does}; (11) RTS: achieving high robust accuracy with semisupervised learning procedure (self-training)~\cite{carmon2019}; (12)~\cite{gowal2020uncovering}: achieving state-of-the-art results by combining larger models, Swish/SiLU activations and model weight averaging. 
 
 In Table~\ref{aa-table}, we show that our proposed method achieves a clear superiority over almost all the defence approaches on both the clean data and adversarial examples on CIFAR-10. One exception is  on clean data:  ours is  slightly inferior to \cite{gowal2020uncovering} which is however trained with additional data. On CIFAR-100, as observed, overall, our proposed method achieves significantly better performance than all the other defence approaches (without data augmentation) both on the clean data and AA attacked examples. Furthermore, it is noted that, while our ATLD+ method is just slightly inferior to \cite{gowal2020uncovering} (which is trained with additional data), it is substantially ahead of the normal version of \cite{gowal2020uncovering}.
 

\subsection{Robustness Evaluation under Adaptive Attacks}
The proposed IMT utilizes information from the discriminator during the inference time, so what if the discriminator is white-box attacked during inference? To verify the effectiveness of the proposed IMT. we further perform additional experiments to evaluate adaptive attacks focusing on both the classifier and the discriminator as shown in Table~\ref{adaptive}. We find that when the adversary attacks both the classifier and discriminator adaptively, the proposed ATLD still has high robustness. Here ‘D’ means the binary classification of the discriminator. ‘Cla’ means the classification loss of the classifier such as the cross entropy loss and CW loss. ‘(Cla+D)-step20’ means that both the loss of and classifier and the discriminator are used to compute the attacking gradients with 20 steps, ‘Cla-step20 + D-step1’ means that the classification loss is first used to compute attacking gradients with 20 steps then followed by 1 step compute attacking gradients with the discriminating loss. Note that FGSM only takes 1 step.

\begin{table}[htbp]
\caption{Robustness Accuracy under Different Adaptive Attacks} 
\label{adaptive}

\begin{center}
\begin{sc}
\scalebox{0.8}
{
\begin{tabular}{lccccccccccccc} 
\toprule
\multicolumn{5}{c}{Attacked Models (CIFAR-10)}\\
\hline
Methods  & (Cla+D)-step20 & Cla-step20 + D-step1 &  D-step1 & D-step20 \\
         & FGSM/PGD/CW & FGSM/PGD/CW & & \\
\hline
ATLD & 84.66/80.4/79.38 & 86.07/81.16/81.13 & 87.51 & 85.72\\
ATLD$+$ & 85.4/80.31/79.49 & 85.72/83.54/83.32 & 87.68 & 87.75\\
\midrule

\multicolumn{5}{c}{Attacked Models (CIFAR-100)}\\
\hline
Methods  & (Cla+D)-step20 & Cla-step20 + D-step1 &  D-step1 & D-step20 \\
         & FGSM/PGD/CW & FGSM/PGD/CW & & \\
\hline
ATLD & 56.91/53.95/50.37 &  57.81/58.59/57.57 & 59.62 & 58.84\\
ATLD$+$ & 57.01/51.60/47.44 & 58.08/60.32/58.62 & 59.28 & 61.50\\

\midrule
\multicolumn{5}{c}{Attacked Models (SVHN)}\\
\hline
Methods  & (Cla+D)-step20 & Cla-step20 + D-step1 &  D-step1 & D-step20 \\
         & FGSM/PGD/CW & FGSM/PGD/CW & & \\
\midrule
ATLD & 89.16/79.64/79.18 & 87.13/83.63/83.19 & 89.75 & 82.85\\
ATLD$+$ & 89.03/78.86/78.14 & 87.31/85.18/85.14 & 90.24 & 85.14\\
\bottomrule

\end{tabular}
}
\end{sc}
\end{center}

\end{table}

\section{Conclusion}
We have developed a novel adversarial training method which leverages both local and global information to defend against adversarial attacks in this paper. In contrast, existing adversarial training methods mainly generate adversarial perturbations in a local and supervised fashion, which could however limit the model's generalization. 
We have established our novel framework via an adversarial game between  a discriminator and a classifier: the discriminator is  learned to differentiate globally the latent distributions of the natural data and the perturbed counterpart, while the classifier is trained to recognize accurately the perturbed examples as well as enforcing the invariance between the two latent distributions. 
Extensive empirical evaluations have shown the effectiveness of our proposed model when compared with the recent state-of-the-art in defending adversarial attacks in both the white-box and black-box settings.


%

\appendices
\section{Detailed Derivation}
In the main content of this paper, we defined our proposed main objective as:

\begin{eqnarray}
\begin{small}
\begin{aligned}
\min_\theta  &\Big\{ \sum_{i=1}^N\underbrace{L(x^{adv}_i,y_i;\theta)}_{L_f} + \\ & \sup_W\sum_{i=1}^N[\underbrace{\log D_W(f_{\theta}(x^{adv}_i))+(1-\log D_W(f_{\theta}(x_i)))}_{L_d}]\Big\} 
\\
\text{s.t.} & \quad x_i^{adv}=\arg \max_{x'_i\in B(x_i,\epsilon)}[\log D_W(f_{\theta}(x'_i))
\\&\qquad\qquad\qquad\qquad\qquad+(1-\log D_W(f_{\theta}(x_i)))]
\label{re_p}
\end{aligned}
\end{small}
\end{eqnarray}

In this section, we provide the details about the derivation for the main objective function Equation.~(\ref{re_p}) (exactly the same as Equation.~(\ref{re_p1}) in the main paper) and elaborate on how to compute the adversarial examples and the transformed examples.

\subsection{Derivation for Main Objective Function~(Equation.~(\ref{re_p}))}

We start with minimizing the largest $f$-divergence between latent distributions $P_\theta$ and $Q_\theta$ induced by perturbed example $x'$ and natural example $x$. And we denote their corresponding probability density functions as $p(z)$ and $q(z)$. According to Equation.~(\ref{f_d}), we have

\begin{eqnarray}
\begin{small}
\begin{aligned}
& \min_\theta \max_{Q_\theta} D_f(P_\theta||Q_\theta) \\&  =\min_\theta \max_{q(z)} \int_{\mathcal{Z}}q(z)\sup_{t\in \dom f^*} \{t \frac{p(z)}{q(z)}-f^*(t)\}dx
\\ &\geq \min_\theta \max_{q(z)} \sup_{T\in \tau}(\int_{\mathcal{Z}}p(z)T(z)dz \\& \qquad\qquad\qquad\qquad -\int_{\mathcal{Z}}q(z)f^*(T(z))dz)
\\ &=\min_\theta \max_{Q_\theta} \sup_W \Big \{\mathbb{E}_{z\sim P_\theta}[g_f(V_W(z))]\\& \qquad\qquad\qquad\qquad+\mathbb{E}_{z\sim Q_\theta}[-f^*(g_f(V_W(z)))]\Big \}
\\&=\min_\theta \sup_W\Big \{\mathbb{E}_{x\sim \mathcal{D}} \big \{\max_{x'\in B(x,\epsilon)}[g_f(V_W(f_\theta(x')))]\\& \qquad\qquad\qquad\qquad+[-f^*(g_f(V_W(f_\theta(x))))]\big \}\Big \}
\label{derive}
\end{aligned}
\end{small}
\end{eqnarray}

To compute the Jensen-Shannon divergence between $P_\theta$ and $Q_\theta$, we set $g_f(t)=-\log(1+e^{-t})$ and $f^*(g)=-\log(2-e^g)$. Then, we have

\begin{eqnarray}
\begin{small}
\begin{aligned}
 &\min_\theta \max_{Q_\theta}  D_{JS}(P_\theta||Q_\theta) \\&\geq\min_\theta \sup_W\Big \{\mathbb{E}_{x\sim \mathcal{D}}\big \{\max_{x'\in B(x,\epsilon)}[\log D_W(f_\theta(x')))]\\&\qquad\qquad\qquad\qquad\qquad+[1-\log D_W(f_\theta(x))))]\big \}\Big \}
\label{derive1}
\end{aligned}
\end{small}
\end{eqnarray}
where $D_W(x) = 1/(1+e^{-V_W(x)})$ is equivalent to optimize the lower bound of Jensen-Shannon divergence between $P_\theta$ and $Q_\theta$. With disentangling the computation of adversarial examples from Eq.~(\ref{derive1}) and further considering the classification loss for the classifier $L_f$ and the discriminator $L^{1:C}_d$, we can obtain the final objective:
\begin{eqnarray}
\begin{small}
\begin{aligned}
\min_\theta& \Big\{ \sup_W\sum_{i=1}^N[\underbrace{\log D^0_W(f_{\theta}(x^{adv}_i))+(1-\log D^0_W(f_{\theta}(x_i))}_{L^0_d}]\\ &+\underbrace{L(x^{adv}_i,y_i;\theta)}_{L_f}\\&+\min_{W}[\underbrace{l(D^{1:C}_W(f_{\theta}(x_i)),y_i)+l(D^{1:C}_W(f_{\theta}(x^{adv}_i)),y_i)]}_{L^{1:C}_d}  \Big\},\\
\text{s.t.} & \quad x_i^{adv}=\arg \max_{x'_i\in B(x_i,\epsilon)}[\log D^0_W(f_{\theta}(x'_i))\\&\qquad\qquad\qquad\qquad\qquad+(1-\log D^0_W(f_{\theta}(x_i))] 
\label{derive2}
\end{aligned}
\end{small}
\end{eqnarray}

\subsection*{Computation for Adversarial Example and Transformed Example}
To compute the adversarial example, we need to solve the following problem:
\begin{eqnarray}
\begin{small}
\begin{aligned}
x_i^{adv}=\arg \max_{x'_i\in B(x_i,\epsilon)}[\underbrace{\log D^0_W(f_{\theta}(x'_i))+(1-\log D^0_W(f_{\theta}(x_i))}_{L^0_d}]
\label{derive3}
\end{aligned}
\end{small}
\end{eqnarray}
It can be reformulated as computing the adversarial perturbation as follows: 
\begin{eqnarray}
\begin{aligned}
r_i^{adv}=\arg \max_{\|r\|_\infty \leq \epsilon}[L^0_d(x_i+r_i,\theta)] 
\label{derive4}
\end{aligned}
\end{eqnarray}
We first consider the more general case $\|r\|_p \leq \epsilon$ and expand~(\ref{derive4}) with the first order Taylor expansion as follows:
\begin{eqnarray}
\begin{aligned}
r_i^{adv}=\arg \max_{\|r\|_p \leq \epsilon}[L^0_d(x_i,\theta)] + \nabla_x\mathcal{F}^T r_i
\label{derive5}
\end{aligned}
\end{eqnarray}
where $\mathcal{F}=L(x_i,\theta)$. The problem~(\ref{derive5})  can be reduced to:
\begin{eqnarray}
\begin{aligned}
\max_{\|r_i\|_p = \epsilon}\nabla_x\mathcal{F}^T r_i
\label{derive6}
\end{aligned}
\end{eqnarray}
We solve it with the Lagrangian multiplier method and we have
      \begin{eqnarray}
        \begin{aligned}
         \nabla_{x}\mathcal{F}r_i = \lambda (\|r_i\|_p-\epsilon)
        \label{derive7}
        \end{aligned}
        \end{eqnarray}
Then we make the first derivative with respect to $r_i$:
      \begin{eqnarray}
        \begin{aligned}
         \nabla_{x}\mathcal{F} = \lambda \frac{r_i^{p-1}}{p(\sum_{j}(r^j_i)^p)^{1-\frac{1}{p}}}
       \label{derive8}
        \end{aligned}
        \end{eqnarray}
      \begin{eqnarray*}
        \begin{aligned}
         \nabla_{x}\mathcal{F} = \frac{\lambda}{p}(\frac{r_i}{\epsilon})^{p-1}
        \label{derive9}
        \end{aligned}
        \end{eqnarray*}
      \begin{eqnarray}
        \begin{aligned}
         (\nabla_{x}\mathcal{F})^{\frac{p}{p-1}} = (\frac{\lambda}{p})^{\frac{p}{p-1}}(\frac{r_i}{\epsilon})^{p}
        \label{derive10}
        \end{aligned}
        \end{eqnarray}
    If we sum over two sides, we have
    \begin{eqnarray}
        \begin{aligned}
         \sum(\nabla_{x}\mathcal{F})^{\frac{p}{p-1}} = \sum(\frac{\lambda}{p})^{\frac{p}{p-1}}(\frac{r_i}{\epsilon})^{p}
        \label{derive11}
        \end{aligned}
        \end{eqnarray}
     \begin{eqnarray}
        \begin{aligned}
         \|\nabla_{x}\mathcal{F}\|_{p^*}^{p^*} = (\frac{\lambda}{p})^{p^*}*1
        \label{derive12}
        \end{aligned}
        \end{eqnarray}
where $p^*$ is the dual of $p$, i.e. $\frac{1}{p}+\frac{1}{p^*}=1$. We have
     \begin{eqnarray}
        \begin{aligned}
          (\frac{\lambda}{p}) = \|\nabla_{x}\mathcal{F}\|_{p^*}
        \label{derive13}
        \end{aligned}
        \end{eqnarray}
By combining (\ref{derive10}) and (\ref{derive13}), we have
        \begin{eqnarray}
        \begin{aligned}
        r^*_i=&\epsilon \mathrm{sgn}(\nabla_x \mathcal{F})(\frac{\lvert \nabla_x \mathcal{F} \rvert}{\|\nabla_x \mathcal{F} \|_{p^*}})^{\frac{1}{p-1}}
        \\ =&\epsilon \mathrm{sgn}(\nabla_x  L^0_d)(\frac{\lvert \nabla_x  L^0_d \rvert}{\|\nabla_x  L^0_d \|_{p^*}})^{\frac{1}{p-1}}
        \label{derive14}
        \end{aligned}
        \end{eqnarray}
In this paper, we set $p$ to $\infty$. Then we have
        \begin{eqnarray}
        \begin{aligned}
        r^*_i &= \epsilon \lim_{p\to \infty}sgn(\nabla_x L^0_d)(\frac{\lvert \nabla_x  L^0_d \rvert}{\|\nabla_x  L^0_d\|_{p^\ast}})^{\frac{1}{p-1}}
        \\ &= \epsilon \mathrm{sgn}(\nabla_x  L^0_d)(\frac{\lvert \nabla_x  L^0_d \rvert}{\|\nabla_x  L^0_d\|_{1}})^{0}
        \\ &= \epsilon \mathrm{sgn}(\nabla_x  L^0_d)
        \label{derive15}
        \end{aligned}
        \end{eqnarray}

Then we can obtain the adversarial example: 
        \begin{eqnarray}
        \begin{aligned}
        x^*_i =  x_i + \epsilon \mathrm{sgn}(\nabla_x  L^0_d)
        \label{derive16}
        \end{aligned}
        \end{eqnarray}
To compute the transformed example, we need to solve the following problem:
\begin{align}
r^*=\arg\min_{\|r\|_\infty\leq \epsilon}\log D^0_W(f_\theta(x+r)).
\label{derive17}
\end{align}
With the similar method, we can easily get the transformed example $x^t$
\begin{align}
x^t = x - \epsilon \mathrm{sgn}(\nabla_x  \log D^0_W).
\label{derive18}
\end{align}

\section*{Acknowledgment}

The work was partially supported by the following: National
Natural Science Foundation of China under no.61876155
and no.61876154; Jiangsu Science and Technology Programme
(Natural Science Foundation of Jiangsu Province)
under no. BE2020006-4B, BK20181189, BK20181190;
Key Program Special Fund in XJTLU under no. KSF-T-06,
KSF-E-26, KSF-A-10, and XJTLU Research Development
Fund under no. RDF-16-02-49.

\ifCLASSOPTIONcaptionsoff
  \newpage
\fi



\bibliographystyle{IEEEtran}
\bibliography{mybib}
%

%

\begin{IEEEbiography}[{\includegraphics[width=1in,height=1.25in,clip,keepaspectratio]{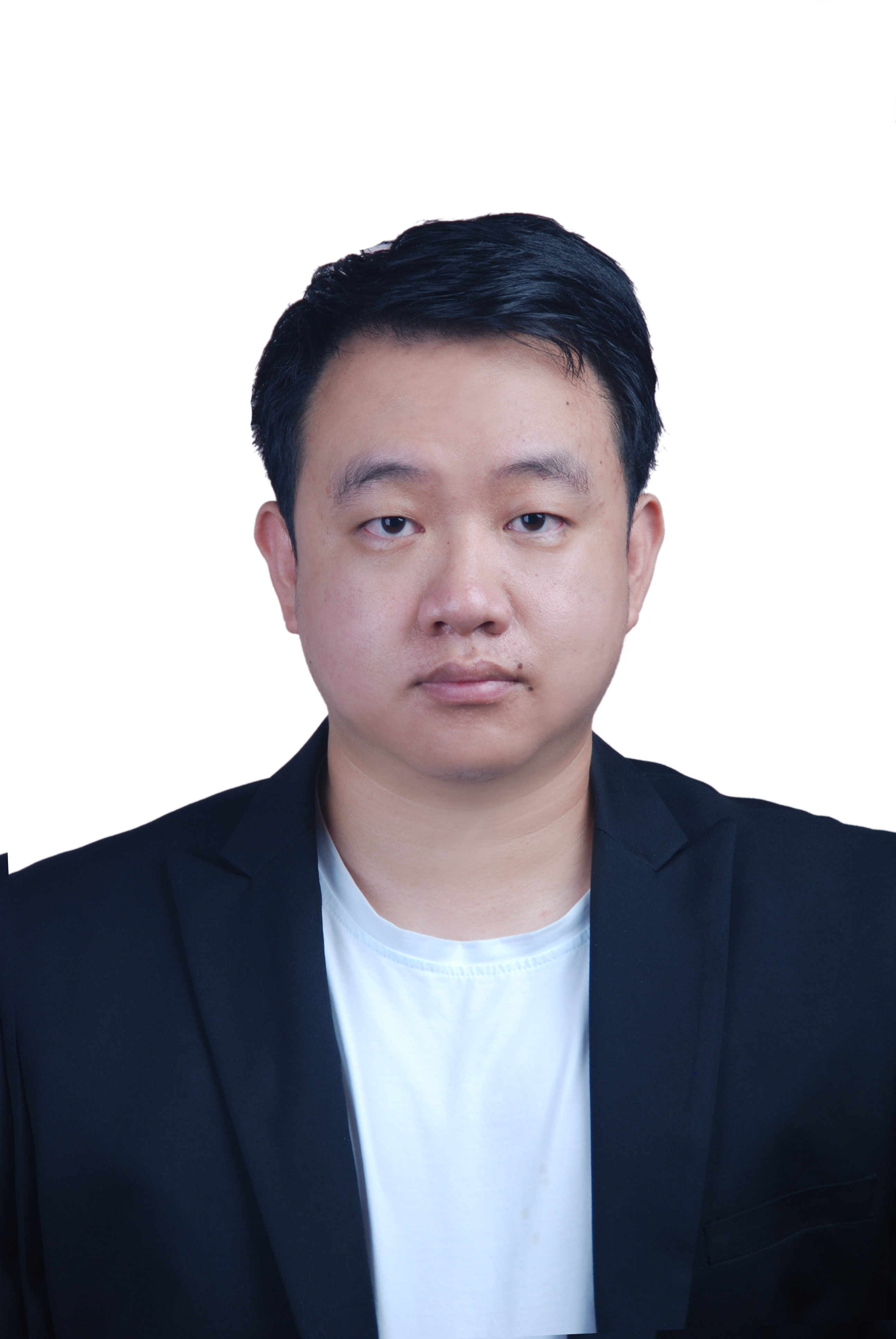}}]{Zhuang QIAN}

	Zhuang QIAN received the B.Eng. degree in Communication engineering Chongqing University of Posts and Telecommunications, Nanan District, Chongqing, P.R.China in 2017. From 2017 $\sim$ now, he is working as a Research Assistant in Department of Electric and Electronic Engineering (EEE Dept.) of the Xian Jiaotong-Liverpool University (XJTLU), Suzhou, Jiangsu, P.R. China. He is currently a Ph.D. student in the School of
	Advanced Technology at Xi’an Jiaotong-Liverpool	University.
	His research interest includes machine learning and deep learning, especially in field classification and robustness.
\end{IEEEbiography}

\begin{IEEEbiography}[{\includegraphics[width=1in,height=1.25in,clip,keepaspectratio]{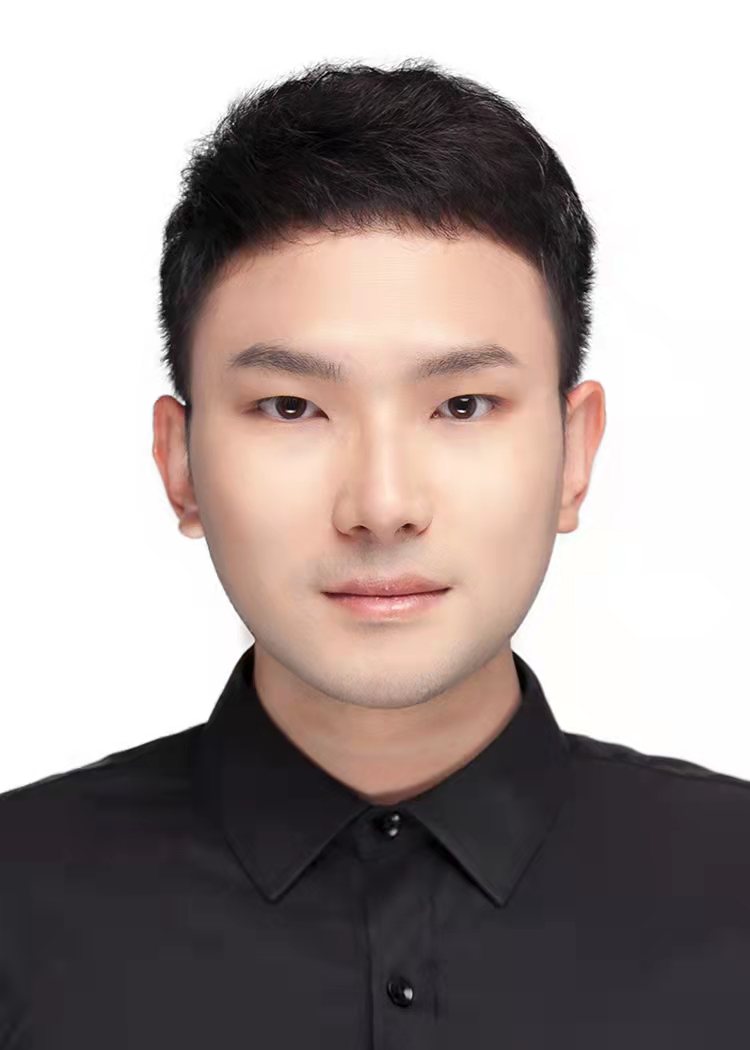}}]{Shufei ZHANG}
Shufei Zhang received the BS from the Electrical and Electronics Engineer department, University of Liverpool, and the MS degrees from the School of Informatics, University of Edinburgh, UK in 2014 and 2015, respectively, and the PhD degree from the Electrical and Electronics Engineer department, University of Liverpool, in 2022. He is currently a Junior Research Fellow with the Shanghai Artificial Intelligence Laboratory, China.
\end{IEEEbiography}


\begin{IEEEbiography}[{\includegraphics[width=1in,height=1.25in,clip,keepaspectratio]{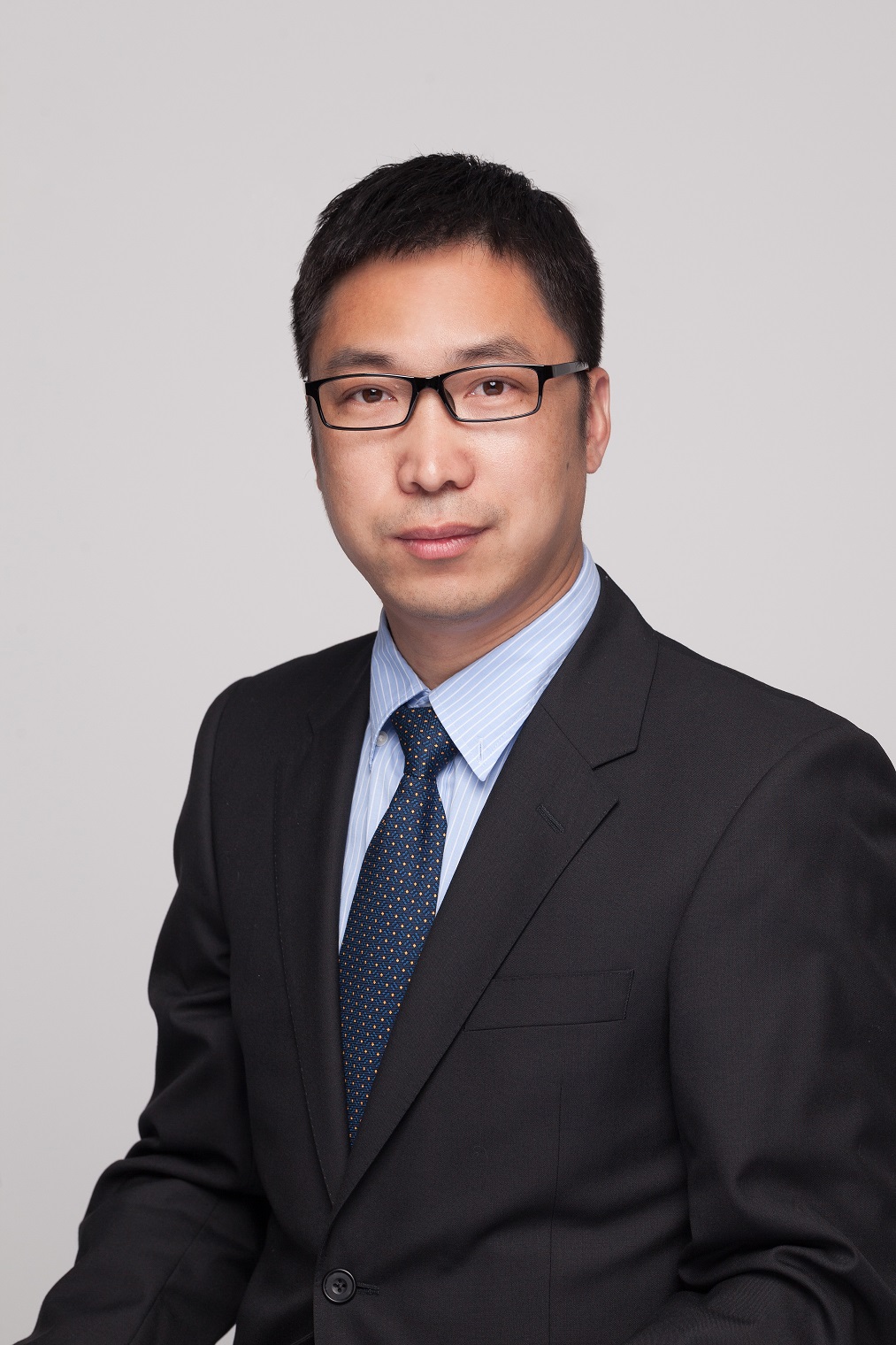}}]{Kaizhu HUANG}
Kaizhu Huang’s research interests include machine learning, neural information processing, and pattern recognition. He is a now a full professor at Institute of Applied Physical Sciences and Engineering, Duke Kunshan University. Before joining DKU, he was a full professor at Department of Intelligent Science, Xi’an Jiaotong-Liverpool University (XJTLU) and Associate Dean of Research in School of Advanced Technology, XJTLU. Prof. Huang obtained his PhD degree from Chinese University of Hong Kong (CUHK) in 2004. He worked in Fujitsu Research Centre, CUHK, University of Bristol, National Laboratory of Pattern Recognition, Chinese Academy of Sciences from 2004 to 2012. He was the recipient of 2011 Asia Pacific Neural Network Society Young Researcher Award. He received best paper or book award six times. He has published 9 books and over 200 international research papers (90+ international journals) e.g., in journals (IEEE T-PAMI, IEEE T-NNLS, IEEE T-BME, IEEE T-Cybernetics, JMLR) and conferences (NeurIPS, ICDM, ICML, IJCAI, SIGIR, UAI, CIKM, ECML, CVPR). He serves as associated editors/advisory board members in 6 journals and book series. He was invited as keynote speaker in more than 30 international conferences or workshops.
\end{IEEEbiography}

\begin{IEEEbiography}[{\includegraphics[width=1in,height=1.25in,clip,keepaspectratio]{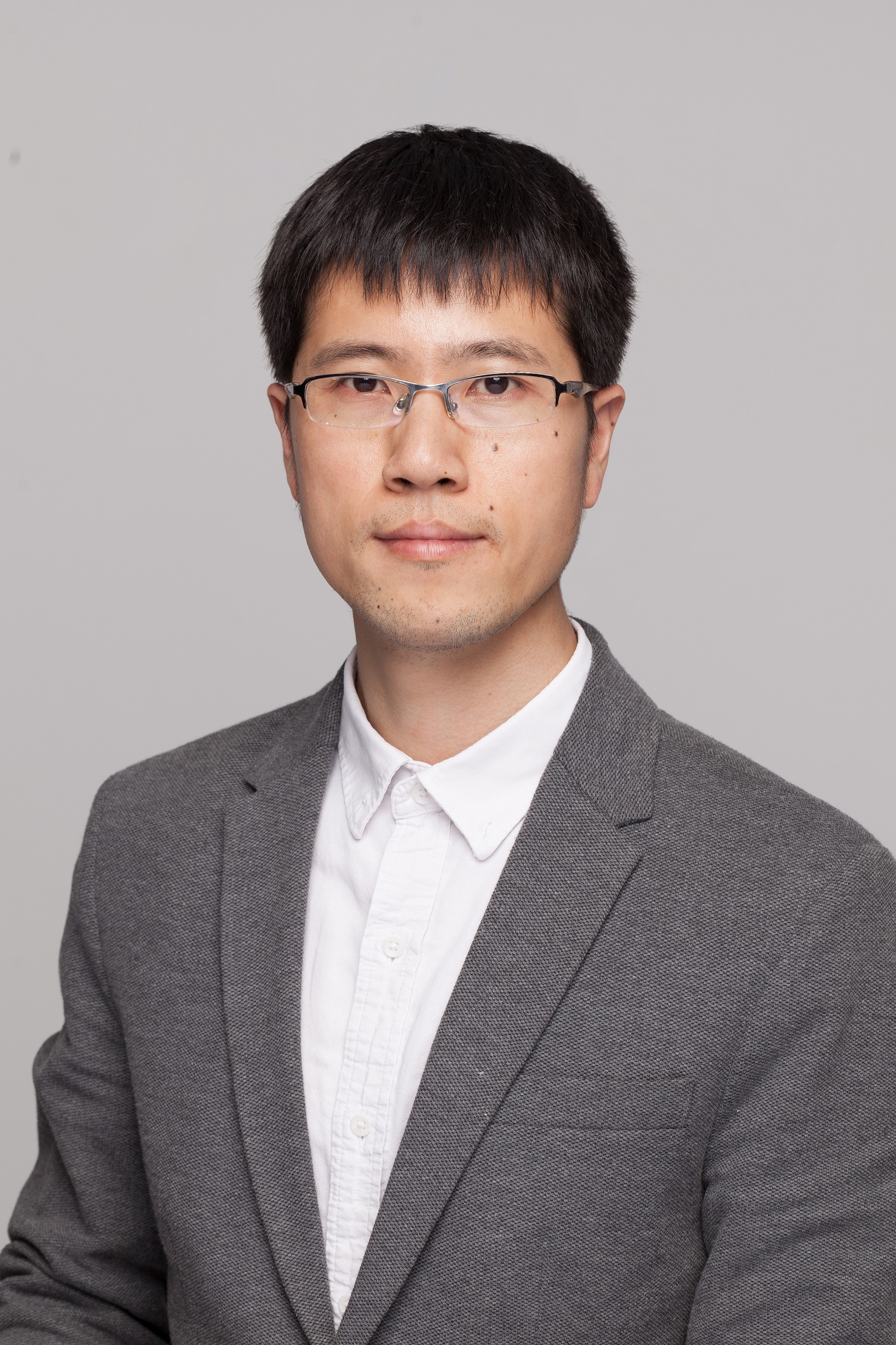}}]{Qiufeng WANG}
Qiu-Feng Wang is currently an associate professor at School of Advanced Technology in Xi’an Jiaotong-Liverpool University (XJTLU). He received the B.Sc. degree in Computer Sicence from Nanjing University of Science and Technology (NJUST) in July 2006, and the Ph.D degree in Pattern Recognition and Intelligence Systems from Institute of Automation, Chinese Academy of Sciences (CASIA) in July 2012. His research interests include pattern recognition and machine learning, more specifically, the document analysis and recognition.
\end{IEEEbiography}

\begin{IEEEbiography}[{\includegraphics[width=1in,height=1.25in,clip,keepaspectratio]{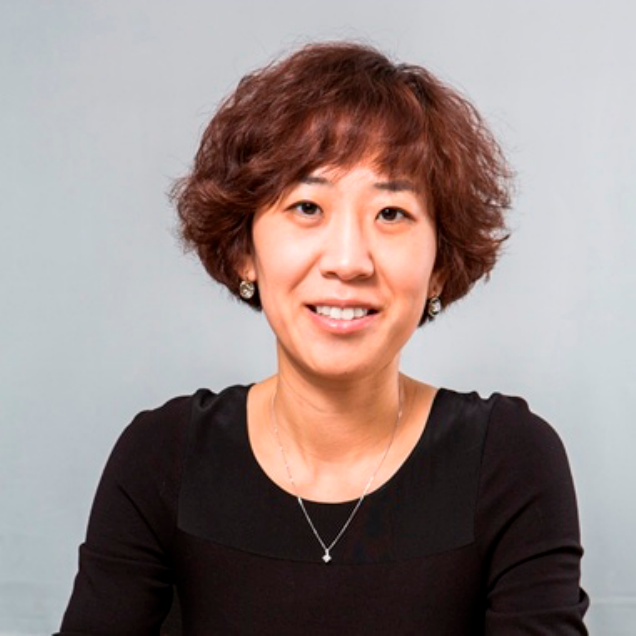}}]{Rui ZHANG}
Rui ZHANG received the First-class (Hons) degree in Telecommunication Engineering from Jilin University of China in 2001 and the Ph.D. degree in Computer Science and Mathematics from University of Ulster, UK in 2007. After finishing her PhD study, she worked as a Research Associate at University of Bradford and University of Bristol in the UK for 5 years. She joined Xian Jiaotong-Liverpool University in 2012 and currently holds the position of Associate Professor. Her research interests include machine learning, data mining and statistical analysis.
\end{IEEEbiography}

\begin{IEEEbiography}[{\includegraphics[width=1in,height=1.25in,clip,keepaspectratio]{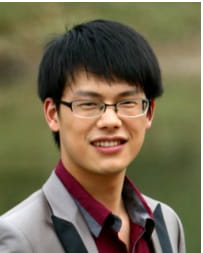}}]{Xinping Yi} (Member, IEEE) received the Ph.D. degree in electronics and communications from T\'el\'ecom ParisTech, Paris, France, in 2015. He is currently a Lecturer (Assistant Professor) with the Department of Electrical Engineering and Electronics, University of Liverpool, U.K. Prior to Liverpool, he was a Research Associate with Technische Universit{\"a}t Berlin, Berlin, Germany, from 2014 to 2017; a Research Assistant with EURECOM, Sophia Antipolis, France, from 2011 to 2014; and a Research Engineer with Huawei Technologies, Shenzhen, China, from 2009 to 2011. His main research interests include information theory, graph theory, machine learning, and their applications in wireless communications and artificial intelligence.
\end{IEEEbiography}




\end{document}